\documentclass{gCMB2e}

\usepackage{subfigure}
\usepackage{url}
\usepackage{algorithm}
\usepackage{algpseudocode}
\usepackage{lmodern}

\begin{document}



\title{Analysis of supervised and semi-supervised GrowCut applied to segmentation of masses in mammography images}

\author{F.R. Cordeiro$^{\rm a}$$^{\ast}$\thanks{$^\ast$Email: frc@cin.ufpe.br\vspace{6pt}}, W.P. Santos$^{\rm b}$$^{\dagger}$\thanks{$^\dagger$Corresponding author. Email:wellington.santos@ieee.org\vspace{6pt}}, and A.G. Silva-Filho$^{\rm a}$$^{\ddagger}$\thanks{$^\ddagger$Email:agsf@cin.ufpe.br
\vspace{6pt}}\\\vspace{6pt} $^{a}${\em{Center for Informatics, Federal University of Pernambuco, Brazil}}\\
$^{b}${\em{Department of Biomedical Engineering, Federal University of Pernambuco, Brazil}}\\\received{v1.0 released January 2015} }

\maketitle

\begin{abstract}
Breast cancer is already one of the most common form of cancer worldwide. Mammography image analysis is still the most effective diagnostic method to promote the early detection of breast cancer. Accurately segmenting tumors in digital mammography images is important to improve diagnosis capabilities of health specialists and avoid misdiagnosis. In this work, we evaluate the feasibility of applying GrowCut to segment regions of tumor and we propose two GrowCut semi-supervised versions. All the analysis was performed by evaluating the application of segmentation techniques to a set of images obtained from the Mini-MIAS mammography image database. GrowCut segmentation was compared to Region Growing, Active Contours, Random Walks and Graph Cut techniques. Experiments showed that GrowCut, when compared to the other techniques, was able to acquire better results for the metrics analyzed. Moreover, the proposed semi-supervised versions of GrowCut was proved to have a clinically satisfactory quality of segmentation.
\end{abstract}


\section{Introduction}
Breast cancer is the most common cancer in woman worldwide, and it is estimated that 1.1 million new cases appear each year, according to the World Health Organization (WHO), described by \cite{mathers2008global}. Breast cancer survival rates can vary between 80\%, in high-income countries, to below 40\% in low-income ones, as cited by \cite{coleman2008cancer}. The low survivability in some countries is related to the lack of early detection programs. Early detection has an important impact on the successful treatment of the cancer, which is hard in late stages. According to \cite{islam2012mammography}, one of the most effective methods for breast cancer analysis is mammography. However, the interpretation of mammographies can be a hard task even to a specialist, since it is affected by the image quality, the radiologist's experience, and the shape of the mammary lesions.

As estimated by the Brazilian National Institute of Cancer \citep{msaude}, the period between the beginning of the tumor and its growth until it becomes palpable, reaching around 1cm, is about 10 years. During this period, breast imaging is essential for tumor monitoring \citep{msaude}. Correct evaluation of the tumor size takes an important role in the planning of the breast cancer treatments, avoiding mutilating surgeries, such as mastectomy, described by \cite{litiere2012breast}. However, imaging devices used by BMH (Brazilian Ministry of Health) for the detection of breast cancer are quite inefficient at the evaluation of the nodule size, and these methods depend substantially on the professional examiner's experience \citep{porto2013aspectos}. Furthermore, digital image diagnosis is complex, mainly because of the large variability of cases. Many cases seen in clinic practice do not fit classic images and descriptions precisely \citep{juhl2000interpretaccao}. For these reasons, mammography computer aided diagnosis (CAD) has been playing an import role to assist radiologists in improving the accuracy of their diagnoses. Consequently, traditional techniques in image processing have been applied in the medical field to make diagnosis less susceptible to errors through accurate identification of anatomic anomalies \citep{da2010algorithm, ye2010medical}.

Recent works, as those of \cite{liu2011new} and \cite{mohamed2009mass}, are quite accurate in identifying the location of tumors. However, little research has been done to verify the quality of segmentation. The shape of the segmented tumor is a determinant factor in the mammogram diagnosis. It is related to the gravity of the tumor and the difference of a few centimeters in the maximum diameter can determine if a surgery is necessary or not. However, it can be very difficult to detect the contour of the tumor accurately depending on several factors, such as shape of the tumor, density, size, location and image quality.

\cite{cordeiro2012segmentation} uses the general purpose user interactive technique GrowCut \citep{vezhnevets2005growcut} applied to segment images of mammography. This technique has been successfully applied in other areas, including medical image segmentation \citep{6412032}. The GrowCut also has the advantage of having no parameters to tune, which usually is not common in segmentation methods. \cite{cordeiro2012segmentation} applies the GrowCut algorithm for the first time to perform mass segmentation. The quality of segmentation is evaluated through metrics of similarity with the ground truth images.

	\textbf{This work improves the method presented by \cite{cordeiro2012segmentation}, where its was presented the use of GrowCut to segment breast tumor, and extends the analysis to state-of-the-art segmentation techniques, proposing two new semi-supervised modifications of GrowCut. The first proposed method combines a multi-layer threshold based approach with morphological operations to generate automatic seeds to the GrowCut, allowing a non-supervised segmentation. The second proposal of this work, called Fuzzy GrowCut, is based on two basic modifications: a) automatic selection of internal points using the differential evolution optimization algorithm, maximizing the minimum distance between these points and the minimum gray level of the associated pixels, in order to minimize the need of human intervention; b) modification of the GrowCut cellular automata evolution rules by introducing Gaussian fuzzy membership functions, in order to make the algorithm able to deal with complex and non-defined mammary lesion borders. This work differs from the previous one contributing to the proposing variations of GrowCut segmentation, which requires less specialist knowledge and which provides segmentation quality equal or better when compared to supervised techniques, regarding metrics that evaluate the similarity of lesion shapes to ground-truth results. The extended results of classical GrowCut, the results of the semi-supervised approach and Fuzzy GrowCut are shown in the Results section}.

This work is organized as following: in section \ref{sec:methods} we present the theoretical background, where the GrowCut algorithm is described, as well as our proposal of semi-supervised GrowCut and Fuzzy GrowCut; in section \ref{sec:methods} we also present the experimental environment, the metrics we used to evaluate our proposal with other techniques, and qualitative comparisons with other segmentation methods; results are presented in section \ref{sec:results}; finally, in sections \ref{sec:discussion} and \ref{sec:conclusion} we present discussions and general conclusions, respectively.

\section{Methods} \label{sec:methods}

In this section are described the implemented techniques, the proposed semi-supervised approaches,  the metrics used for the evaluation and the experimental environment.

\subsection{The GrowCut Algorithm}

GrowCut is a user interactive methodology based on cellular automaton (CA), defined by \cite{hernandez1996cellular}, being used to perform image processing tasks, such as noise reduction, edge and morphological detection, as applied by \cite{rosin2010image}. The CA consists of a grid of cells, in which each cell can assume one of a finite number of states, which vary according to neighborhood rules. The neighborhood corresponds to the selection of neighbor pixels in an image, a process which can use mathematical models such as Neumann and Moore neighborhood, as described by \cite{vezhnevets2005growcut}. All cells update their states according to the same update rule, based on the values of their neighbor cells. Each time the rules are applied to the grid, a new generation is started. 

\begin{figure}[!htb]
	\centering
	\includegraphics[width= 0.8\textwidth]{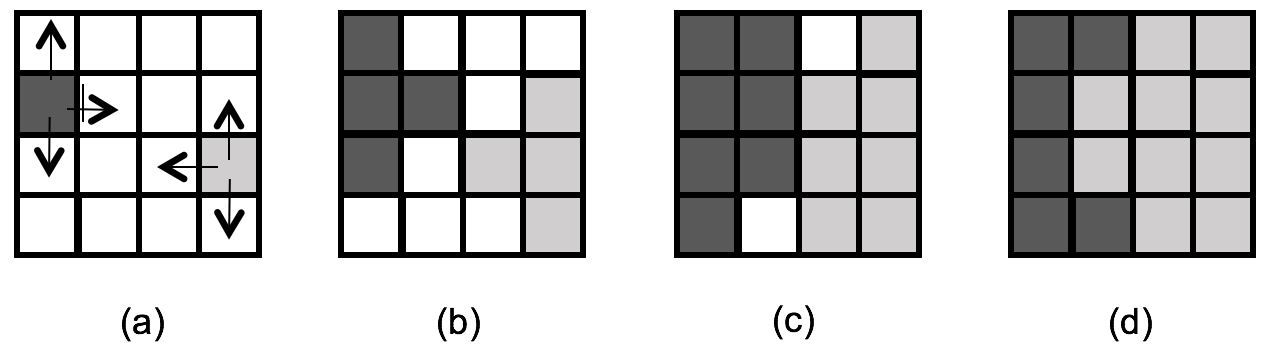}
	\caption{GrowCut's cells evolution}
	\label{fig:cell_ev}
\end{figure}

The GrowCut technique uses the concept of seed pixels, in which the user initially labels a set of pixels, called seeds, and based on those seeds, the algorithm tries to label all pixels of the image. An interesting characteristic of GrowCut algorithm is that is has no parameters to tune, which is not common in segmentation algorithms, being based only on the initial position of the seeds. 
In GrowCut, each cell has a strength value and, at each iteration, the cell's neighbor tries to dominate it. If the defender cell has a higher strength than the dominators, then it keeps its label's value. Otherwise the cell inherits the dominators label. The process continues until the algorithm converges and the cells cease to change. Figure \ref{fig:cell_ev} shows the evolution of a cell's domain in the GrowCut technique.

Figure \ref{fig:cell_ev}(a) shows the initial step, with two initial seed pixels. Each seed pixel tries to dominate its neighbor cells. \textbf{Each seed point must have the feature of being close to the edges of the desired area of segmentation. Besides that, the seeds have the feature of being labelled as a seed of background or foreground. Therefore, the correct seed positioning is important to the GrowCut. The example shows only two seed points, one of each label, just for simplification, but in the algorithm more seeds are used.}
 Pixels with no labels have null strength, represented as the white cells. Figure \ref{fig:cell_ev} (b) and (c) shows the spread of the seeds through the pixels with no labels, which have zero strength. In Figure \ref{fig:cell_ev} (c), it shows a situation in which cells from one label try to dominate cells from other labels. The success of the domination is based on the neighborhood strength. 
 This process continues until the algorithm converges, which is represented in Figure \ref{fig:cell_ev} (d). 

The pseudocode of GrowCut is described in Algorithm \ref{alg:growcut}. For each cell $p$ in a $P$ space of cells, the previous states are copied, updating the label value of cell $p$ in iteration $t+1$, represented as $l_{p}^{t+1}$, and the strength value of cell $p$ in iteration $t+1$, as $\Theta_{p}^{t+1}$. Subsequently, for each cell $q$ belonging to the neighborhood of cell $p$, represented by $N(p)$, the update label condition is checked. \textbf{In condition of line 5, $\vec{C_p}$ and $\vec{C_q}$ are tridimensional vectors of cells $p$ and $q$, respectively, and $\Theta_{q}^{t}$ and $\Theta_{p}^{t}$ are values of strength of cells $q$ and $p$ in the iteration $t$. In the original GrowCut, which was created to segmented colored images, the $\vec{C_p}$ is the vector of intensity values of  cell \emph{p} in RGB scale. This shows that the GrowCut also is able to segmented images in RGB scale and can be applied to segment colored medical images. We maintained the original notation to generalize the potential of GrowCut, but for mammograms images, whose images are in grayscale, we simplified it and used the gray level intensity of the cell instead of the tridimensional vector}. The function $g$, in lines 5 and 7, is a decreasing monotonic function, represented by Equation \ref{eq:g}. The \emph{g} function is responsible for adjust the strength of the updated cell based on the difference of intensity values between neighbor cells. \textbf{The range of values of the independent variable x is into the interval [0,255], because it is the range of difference of intensity values between two pixels. The \emph{g} function is the function used in the classical GrowCut and maintained in the proposed approaches. However, an analysis using different functions can be done in future works.}

\begin{algorithm}
\caption{GrowCut evolution rule}\label{alg:growcut}
\begin{algorithmic}[1]
  \ForAll {$p \in P$}
  \State $l_{p}^{t+1} \gets l_p^{t}$ 
  \State $\Theta _{p}^{t+1} \gets \Theta_p^{t}$
 
  \ForAll {$q \in N(p)$}
      \If {$g(\left \| \vec{C_p} - \vec{C_q} \right \|_{2})\cdot \Theta_{q}^{t}>\Theta_{p}^{t}$} 
          \State  $l_{p}^{t+1} \gets l_{q}^{t}$
          \State  $\Theta_{p}^{t+1} \gets g(\left \| \vec{C_p} - \vec{C_q} \right \|_{2})\cdot \Theta_{q}^{t}$
      \EndIf
  \EndFor
 \EndFor
\end{algorithmic}
\end{algorithm}

\begin{equation}
g(x) = 1 - \frac{x}{\max\left \|\vec{C}  \right \|_2}
\label{eq:g}
\end{equation}

Finally, the label and strength of cells are updated in case the domination rule is satisfied, and the process repeats until the algorithm converges.

\subsection{Proposal 1: Semi-Supervised GrowCut}

The Semi-Supervised GrowCut (SSGC) technique is a new proposal developed in this work to avoid the user interaction needed in the selection of seeds to perform a good segmentation. In this new approach, the automatically segmentation is performed from an initial selection of a region of interest and there is no need to identify seed points to perform the segmentation. The manual selection of seeds is not desired because it depends on the specialist knowledge, which can vary depending on the user experience. Therefore, the elimination of the step of selection of seeds is important to the process of segmentation became less dependent on the user. Naturally, a semi-supervised technique usually loss in accuracy compared to the supervised ones, but it is important to have techniques less dependent on the knowledge of the problem and able to perform good segmentation.

\begin{figure}[htp]
\centering
\subfigure[Region of Interest]{
  \includegraphics[width=4cm, height=4cm]{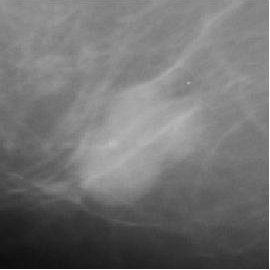}
}
\subfigure[Multilevel Threshold]{
  \includegraphics[width=4cm, height=4cm]{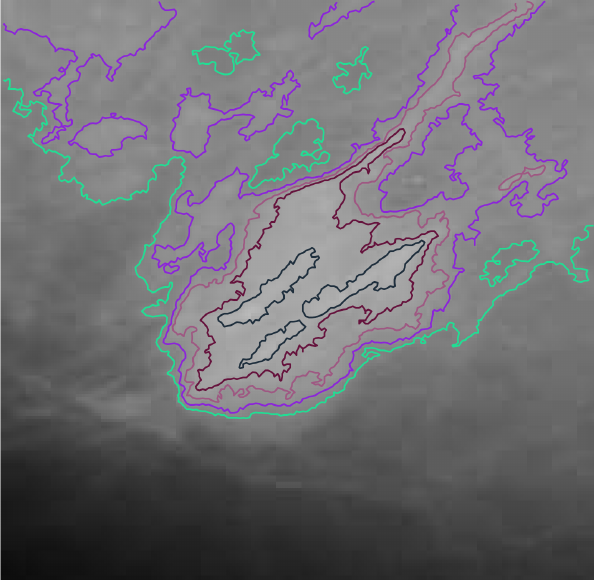}
}
  \caption{Multilevel threshold applied to a Region of Interest.}
  \label{fig:multi}
\end{figure}

The SSGC is a hybrid technique based on GrowCut and multilayer threshold \citep{gao2010combining}. The multilayer threshold (also known as multilevel threshold and multi-threshold) is a technique applied to identify tumor regions in medical images. In this method, initially the mammography image is submitted to different levels of threshold, aiming to identify concentric regions. One of the features identified is that usually the tumor is located in concentric regions, as shown in Figure \ref{fig:multi}.

\textbf{The parameters of multilevel threshold were defined based on the empirical experimentation performed in section \ref{sec:results}.}

\begin{figure}[!htb]
\centering
  \fbox{\includegraphics[width=7.5cm]{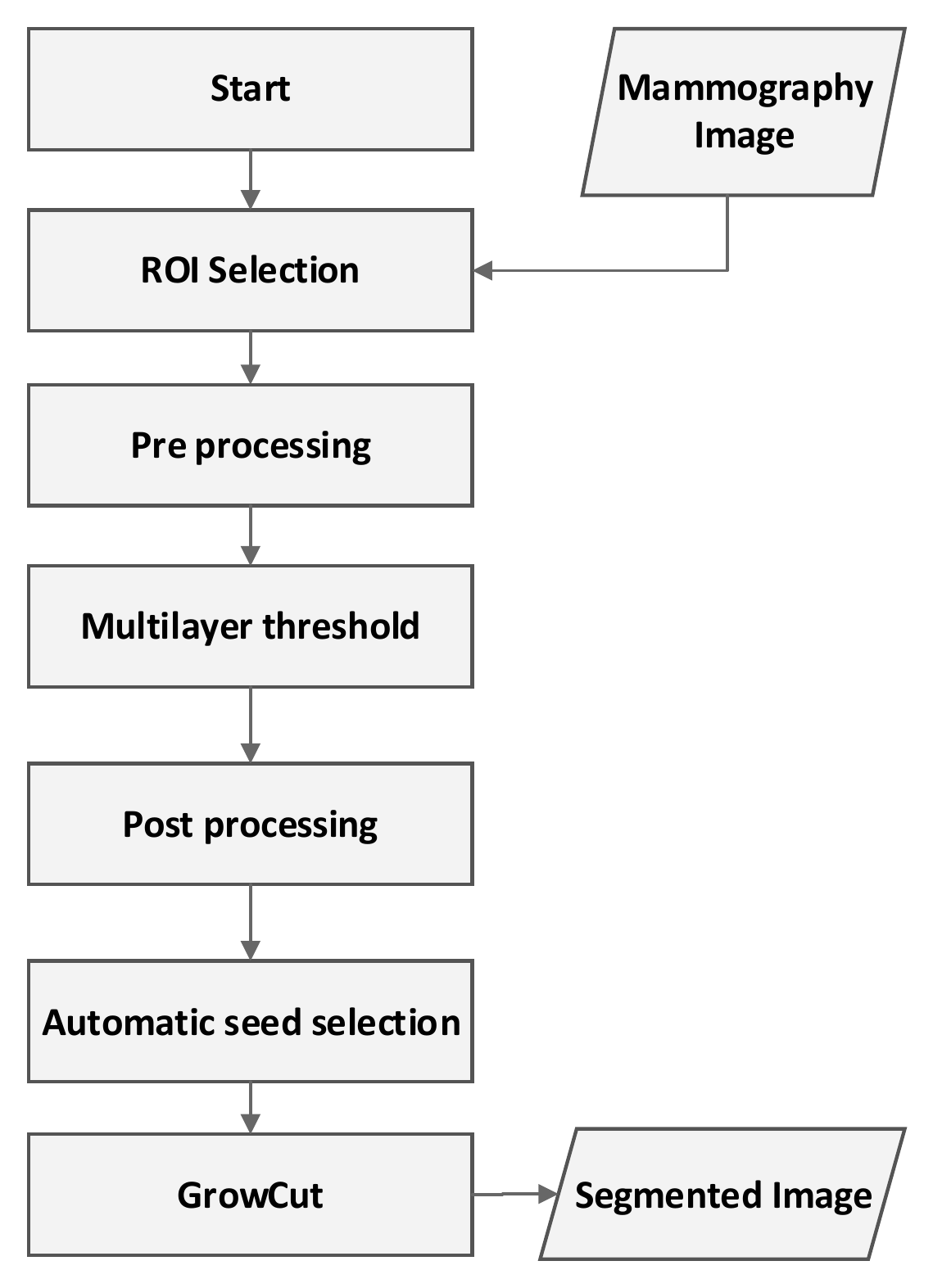}}
  \caption{Semi-Supervised GrowCut Flowchart}
  \label{fig:flowUGC}
\end{figure}

 When applying multilevel threshold, different layers are obtained and the layers with concentric regions are considered as mass regions, while the regions without concentric layers are discarded. After this, a pre-segmentation of the tumor region is obtained. However, the area obtained from the pre-segmentation is not well segmented and therefore GrowCut is applied to improve the contour segmentation. The flowchart of the semi-supervised GrowCut is shown in Figure \ref{fig:flowUGC}.

\begin{figure}[htp]
\centering
  \fbox{\includegraphics[width=9.0cm]{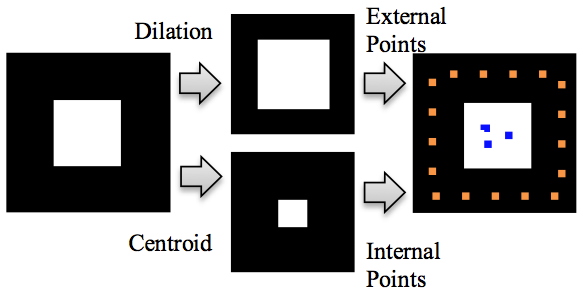}}
  \caption{Automatic selection of internal and external seed pixels}
  \label{fig:seeds}
\end{figure}

As illustrated in Figure \ref{fig:flowUGC}, the first step of the SSGC is the selection of the region of interest by the user. In the proposed work, it was observed that SSGC works better when the ROI is not too large. \textbf{Next, in the pre-processing step, the edge-enhancing anisotropic diffusion filter developed by \cite{weickert1997review} is applied in order to keep the edges sharp. The diffusion process transforms a mammogram to a smooth image from which more informative isocontours can be obtained. The noise removal filter is the same used in the work of \cite{gao2010combining}. After the pre-processing, the multilayer threshold is applied to automatically identify the approximated area of tumor.} Next, a post processing is done in the resulting segmentation to eliminate noise. The post processing aims to eliminate small regions that are generated after multilevel threshold. It calculates the size of all regions in the segmented image and maintains only the region with the major area in the ROI. After that, the process continues by applying the GrowCut to the resulting image. As described in the previous subsection, in the original GrowCut the user is required to select the seed pixels to initiate the segmentation. In that step, the user is required to identify the internal and the external points of the area of interest to be segmented. In the SSGC, this selection of seeds is done automatically. To select automatically the seeds from the post processed image, the algorithm applied the morphologic operation of dilation to identify the external point and calculate the centroid in order to identify the internal points. The result of dilation is a background region mask. However, as the original GrowCut works with seeds, all the points of the region of background were mapped to seeds points. The internal seed points do not use de total internal region, but points close to the centroid of internal region. Using centroid reduces the chance of a wrong seed positioning. After the seed pixels have been selected automatically, the GrowCut is started, which provides a segmented image. The point selection process can be visualized in Figure \ref{fig:seeds}.

 The quality of the segmentation depends on the selection of the seed pixels. As the selection is performed automatically, it is natural that it is not so well positioned as a manual selection. However, the automatic process provides a good quality of segmentation, similar to the manual selection, which is shown in the results section.
 
 \textbf{In this work we used the GrowCut code implementation of \cite{code_growcut}.}

\subsection{Proposal 2:Fuzzy GrowCut}  \label{sec:fgc}

In GrowCut, as in the majority of seed based techniques, the quality of segmentation depends directly on the positions of the initial seeds. Therefore, it depends on the user's knowledge  to select appropriately seeds next to the edge of the object to be segmented. In case some seeds are initially labeled incorrectly, the algorithm may perform an undesired poor segmentation.

	We proposed a modified GrowCut algorithm, which aims to reduce the need for initial specialist knowledge about the contour of the object of interest by reducing the effort of selection of seeds. Moreover, the proposed algorithm aims to be fault tolerant, being able to recover from incorrect seed selection.

In classical GrowCut, all the initial seeds selected by the user have maximum strength value, assigning high weights to seeds with incorrect labels. Unlike classical GrowCut, our modified GrowCut is based on the selection of seeds of only one class: the object of interest. We discard the selection of a background class because, from the seeds of the object of interest, we can estimate a frontier region separating object and background. Therefore, instead of assigning all the labeled cells with maximum strength, all the cells are initialized with zero strength, except the cell corresponding to the center of mass of input seeds. Hence, we assign the maximum value to the cell of the center of mass, once we assume it has a higher chance of being correct labeled. The initialization is performed according to the expressions of Equation \ref{eq:init}.

\begin{equation}
\forall p \in P,  l_{p} = 0, \Theta_{p} = 0, l_{cm} = l_{obj}, \Theta _{cm} = 1,
\label{eq:init}
\end{equation}
where \emph{p} is a cell in space \emph{P} of cells, and $l_{p}$ and $\Theta_{p}$ are the labels and strengths of cell \emph{p}, respectively. Label and strength of the cell of the center of mass of the seeds are represented by $l_{cm}$ and $\Theta_{cm}$, respectively.

In our proposal, we also modified the update rule of GrowCut cells in a way that the attack of each cell is based in a region modeled by a Gaussian fuzzy membership function. The strength of the model will be equal to 1 if the pertinence of a determined cell to the background is higher than the pertinence of the same cell to the object. Otherwise, the strength of the model assumes the strength of the current cell. The update evolution rule of the GrowCut algorithm we propose is shown by the pseudo-code of Algorithm \ref{proposed}, where $\Theta_{M,p}^{t}$ and $\Theta_{M,q}^{t}$ are the model strengths for cells \emph{p} and \emph{q}, respectively, being represented by Equations \ref{eq:mi} to \ref{eq:probobj}, as following:

\begin{algorithm}[t]
\caption{Modified GrowCut evolution rule}\label{proposed}
\begin{algorithmic}[1]
\Procedure{ModifiedGrowCut}{$P,l$} 
  	\State $l_{cm} = l_{obj}$
  	\State $\Theta_{cm} = 1$
  \ForAll {$p \in P$}
  \State $l_{p}^{t+1} \gets l_p^{t}$ 
  \State $\Theta _{p}^{t+1} \gets \Theta_p^{t}$
	\State Calculate $\Theta_{M,p}^t$
  \ForAll {$q \in N(p)$}
			\State Calculate $\Theta_{M,q}^t$
      \If {$g(\left \| \vec{C_p} - \vec{C_q} \right \|_{2})\cdot \Theta_{M,q}^t>\Theta_{M,p}^{t}$} 
          \State Calculate $l_{M,p,q}^t$
					\State  $l_{p}^{t+1} \gets l_{M,p,q}^{t}$
          \State  $\Theta_{p}^{t+1} \gets g(\left \| \vec{C_p} - \vec{C_q} \right \|_{2})\cdot \Theta_{M,q}^{t}$
      \EndIf
  \EndFor
 \EndFor
  \State \textbf{return} $l$
\EndProcedure
\end{algorithmic}
\end{algorithm}

\begin{equation}
\Theta_{M,i}=\left\{
\begin{array}{ll}
1, & {\mu_\mathrm{Bkg} (i) > \mu_\mathrm{Obj} (i)} \\ 
\Theta_i, & {\mu_\mathrm{Bkg} (i) \leq \mu_\mathrm{Obj} (i)}\\
\end{array}
\right.,
\label{eq:mi}
\end{equation}

\begin{equation}
\mu_\mathrm{Bkg} (i) = 1 - \mu_\mathrm{Obj} (i),
\label{eq:probbg}
\end{equation}

\begin{equation}
\mu_\mathrm{Obj} (i) = \exp\left(-\frac{(x_i-x_m)^{2}}{2 \alpha_x  s_x^{2} }\right) \exp\left(-\frac{(y_i-y_m)^{2}}{2 \alpha_y s_y^{2} }\right),
\label{eq:probobj}
\end{equation}

where $\mu_\mathrm{Bkg} (i)$ is the the fuzzy membership degree associated to the uncertainty of the $i$-th cell belongs to the image background, whilst $\mu_\mathrm{Obj} (i)$ is the the fuzzy membership degree associated to the uncertainty of the $i$-th belongs to the object of interest. These fuzzy membership functions are Gaussian functions whose variables {$x_i$} and {$y_i$} correspond to the coordinates of the $i$-th cell in the grid, whereas {$x_m$} and {$y_m$} are the coordinates of the center of mass for the initially selected seeds; $s_x$ and $s_y$ are the standard deviation of initial points, whilst $\alpha_x$ and $\alpha_y$ are the weights of tuning of the Gaussian function, empirically determined according to the problem of interest. The $s_x$ and $s_y$ are obtained calculating the standard deviation of the position of the points generated by the Differential Evolution algorithm. \textbf{In general, a cluster not always can be considered a Gaussian-like function. However, in the proposed approach applied to mass segmentation, it can be assumed that there are always two clusters: background and foreground. Three or more clusters were not considered to this model. We considered a Gaussian-like function aiming simplification of the model. Other approaches in literature, such as Expectation Maximization (EM) \citep{moon1996expectation} and Radial Basis Function (RBF) networks \citep{bors2001introduction}, also assume a Gaussian model.} 

The label of each $p$-th cell, $l_{M,p,q}$, is updated to the background label, represented as $l_{Bkg}$, or the label of $p$-th cell, according to the following expression of Equation \ref{eq:lmq}.
\begin{equation}
l_{M,p,q}=\left\{
\begin{array}{ll}
l_{Bkg}, & {\mu_\mathrm{Bkg} (q) > \mu_\mathrm{Obj} (q)} \\ 
l_{q}, & {\mu_\mathrm{Bkg} (q) \leq \mu_\mathrm{Obj} (q)}\\
\end{array}\right..
\label{eq:lmq}
\end{equation}

{\bfseries 
Next, we show a step-by-step numerical example of the mechanism of the Fuzzy GrowCut. The example uses a 5x5 array, which represents the image to be segmented. Figure \ref{fig:passo_e1} (a) shows the example image, in grayscale, where the white region represents the region of tumor that we want to segment and the gray pixels are the background. Figure \ref{fig:passo_e1} (b) is the corresponding array of pixels intensity values of Figure \ref{fig:passo_e1} (a).
\begin{figure}[htp]
\centering
  \includegraphics[width=9cm]{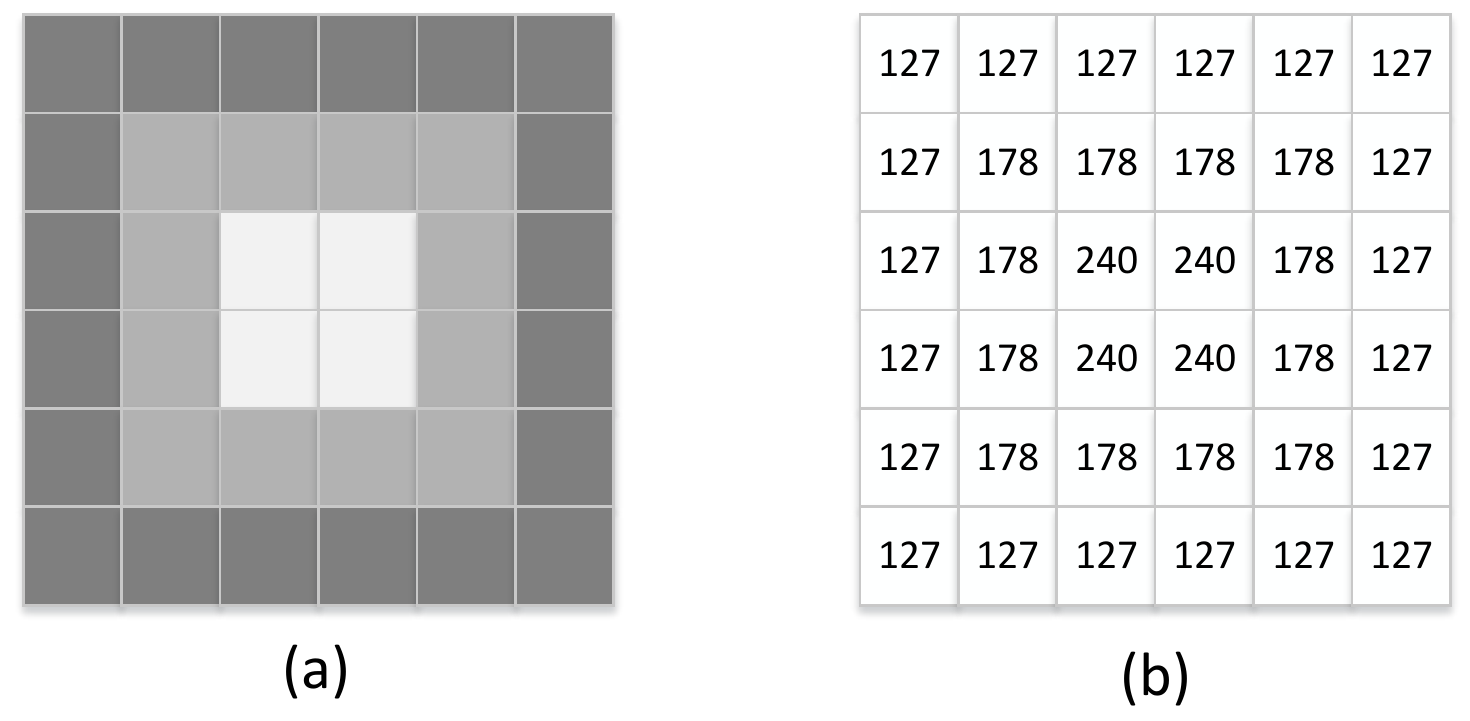}
  \caption{Input image for the step-by-step example of the Fuzzy GrowCut algorithm: (a) Example representing a region of tumor, in grayscale;(b) array of pixels intensity values.}
  \label{fig:passo_e1}
\end{figure}

The first step of the algorithm is to select the seeds inside the region of interest. Figure \ref{fig:passo_e2} (a) shows a selection of seeds, represented as the red circles. As described above, the white region represents the region of tumor. As the proposed approach is tolerant to wrong seed positioning, one seed is set outside the region of tumor, just for demonstration. Next, we calculate the center of mass based on the position of the seeds, shown in Figure \ref{fig:passo_e2} (b). As described in the lines 2 and 3 of algorithm \ref{proposed}, the label of the center of the mass is set to foreground and its strength is set to 1, as shown in Figure \ref{fig:passo_e2}(c), which represents the array of strengths of the the cells. In the initial step all other cells have strength equal to zero and are unlabelled.

\begin{figure}[htp]
\centering
  \includegraphics[width=13cm]{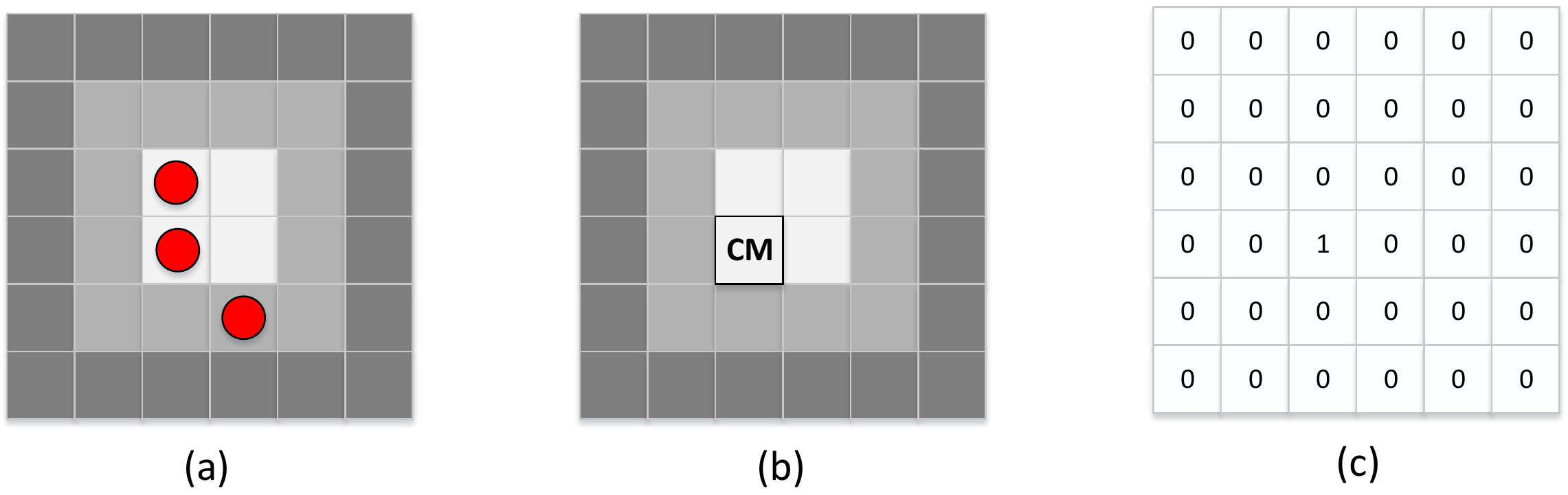}
  \caption{Initialization of the algorithm: (a) seeds selection; (b) center of mass of the seeds; (d) initial strength of each cell.}
  \label{fig:passo_e2}
\end{figure}

Each cell of the array will attack its neighbor based on a fuzzy gaussian membership region. The fuzzy membership degree is calculated based on equations \ref{eq:probbg} and \ref{eq:probobj}. If we use $\alpha$ value of 4 we obtain the membership degrees represented in Figure \ref{fig:passo_e3} (a). In Figure \ref{fig:passo_e3} (a), each cell is divided in two regions, indicating the value of membership to each class: foreground, represented as the white triangle, and background, represented as the gray triangle. These membership degrees will guide the initial strengths and labels, as described in equations \ref{eq:mi} and \ref{eq:lmq}. Based on equation \ref{eq:mi}, Figure \ref{fig:passo_e3} (b) shows the initial strength values obtained, according to the membership degree values. Figure \ref{fig:passo_e3} (c) shows the initial labels, based on equation \ref{eq:lmq} and membership degree values, where the black cells represents the background label, the white cell is the foreground label, and the gray cells are the cells without initial label. 

\begin{figure}[htp]
\centering
  \includegraphics[width=14.0cm]{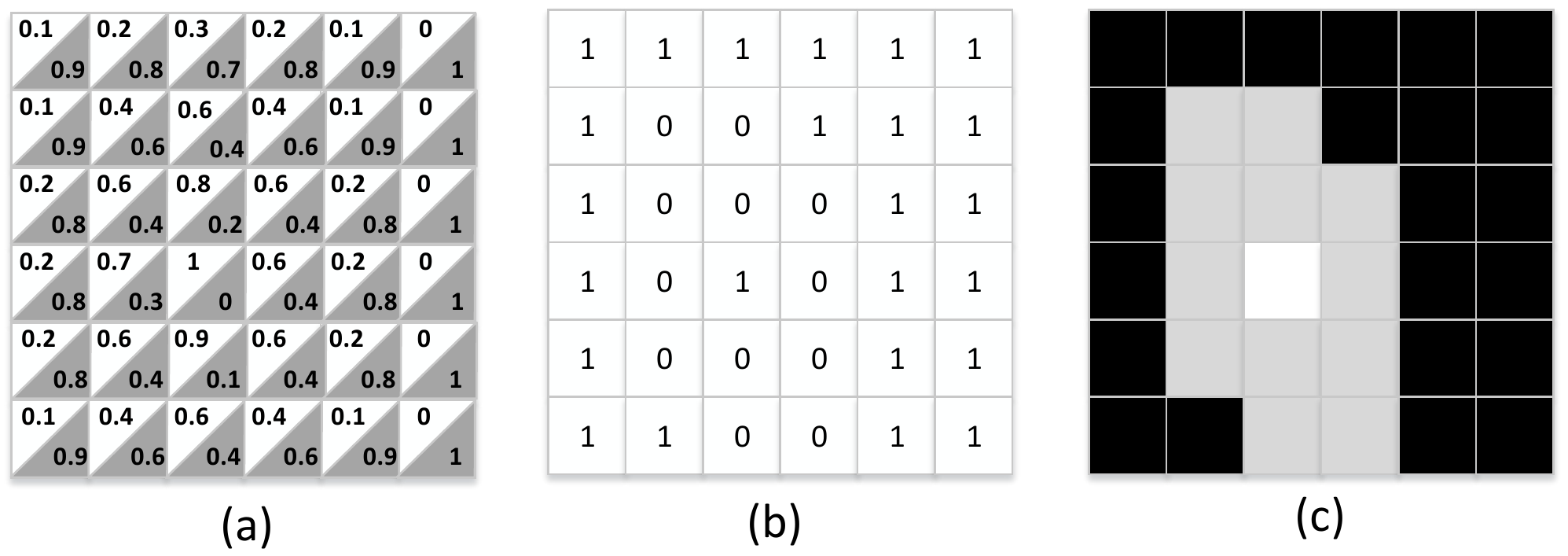}
  \caption{(a) fuzzy-gaussian membership degree values; (b) initial strength of the cells after calculating the membership function degrees; (c) initial labels after calculating the membership function degrees.}
  \label{fig:passo_e3}
\end{figure}

At each iteration of the method, each cell is attacked by its 8 neighbors. Figure \ref{fig:passo_e4} (a) shows a cell being attacked. In this process, the strength of the neighbor and the difference of intensity values between cells will define the winning attacking cell, as defined in line 10 of algorithm \ref{proposed}. The strength of the attacked cell will be updated based on the strength of the winner attacker, as represented by Figure \ref{fig:passo_e4} (b), where the red arrow indicates the winning attacking cell. The label of the cell that is being attacked also is updated based on the attacker cell, as defined in line 12 of algorithm \ref{proposed} and illustrated in Figure \ref{fig:passo_e4} (c) and Figure \ref{fig:passo_e4} (d).

\begin{figure}[htp]
\centering
  \includegraphics[width=\textwidth]{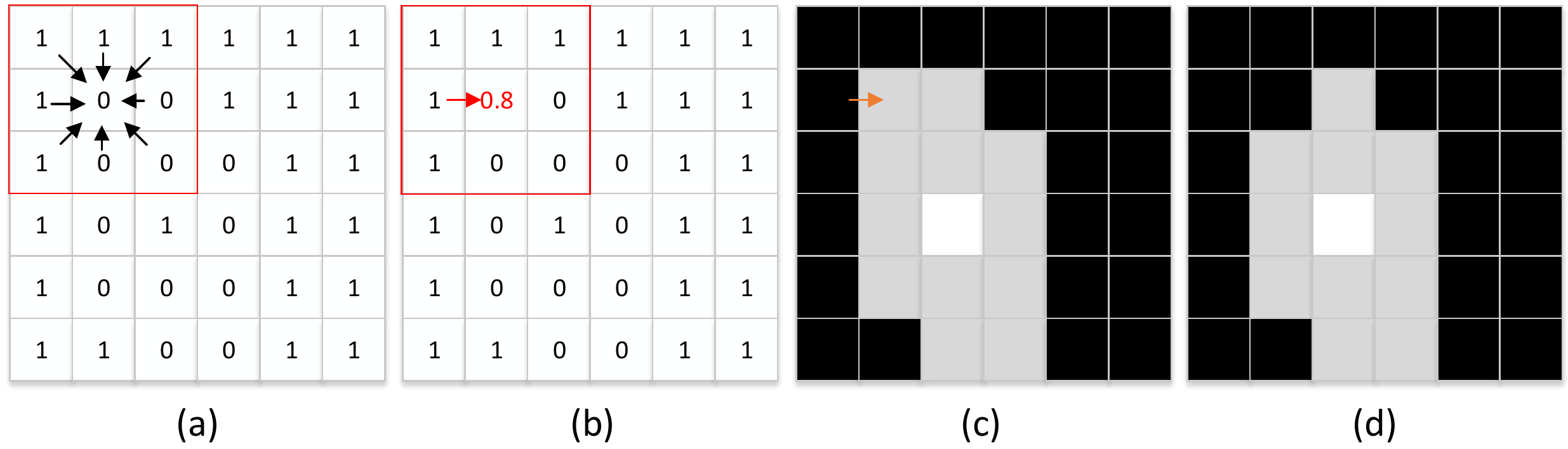}
  \caption{(a) cell being attacked by its neighbors; (b) strength update based on the winning attacking cell; (c) label being updated based on the winning attacking cell; (d) updated label.}
  \label{fig:passo_e4}
\end{figure}

The process of a cell being attacked by its neighbors is performed to all cells of the grid. When all the cells have being attacked and updated its strengths and labels the iteration is finished. Figure \ref{fig:passo_e5} (a) shows the result of strength update at the end of the first iteration. The red values are the updated strengths and the arrows indicates the cells that influenced in the update process. Figure  \ref{fig:passo_e5} (b) shows the update of labels at the end of iteration and the arrows shows the cells that influenced in the update process. Finally, Figure \ref{fig:passo_e5} (c) shows the final segmentation at the end of iteration. The algorithms continues until no cell changes its labels at the end of the iteration. In this example, it was a simple case study and was possible to obtain the segmentation with a single iteration, but in more complex images can be required more iterations.

\begin{figure}[htp]
\centering
  \includegraphics[width=14.0cm]{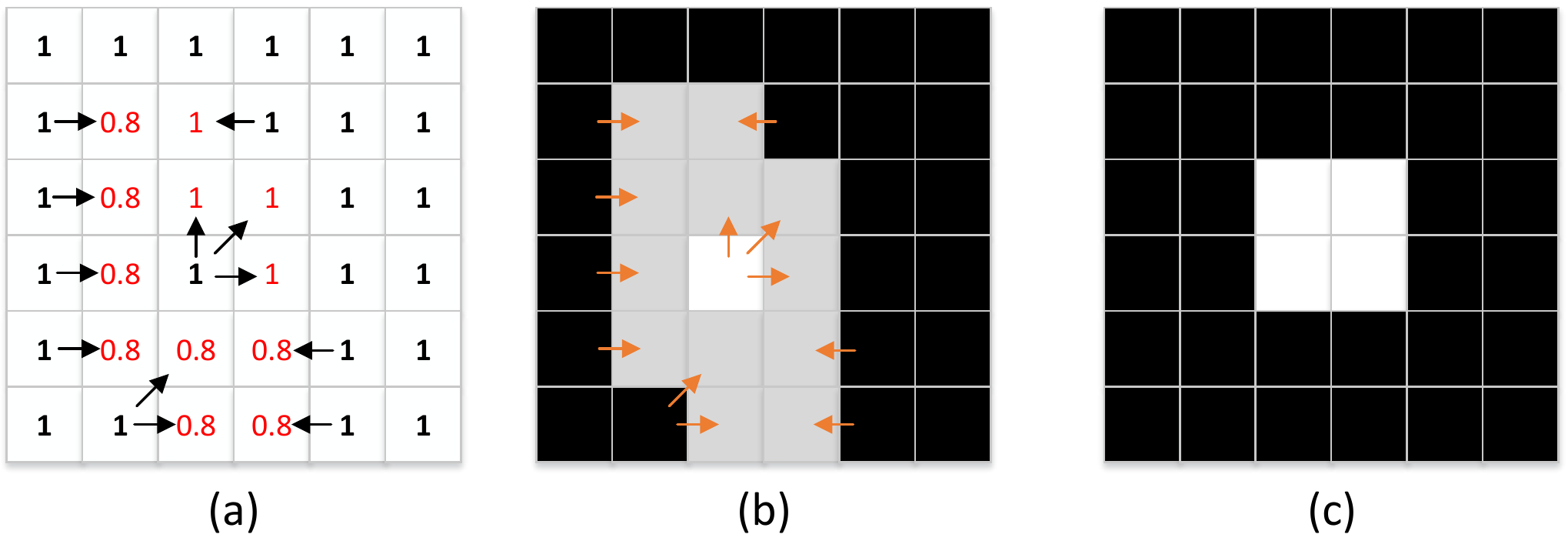}
  \caption{Automatic selection of internal and external seed pixels}
  \label{fig:passo_e5}
\end{figure}

}

Table \ref{comparison} presents a comparison between the classical version and our modified GrowCut algorithm.

\begin{table}[htb]
\footnotesize
\centering
\caption{Comparison between GrowCut and the modified Fuzzy GrowCut Algorithm.}
\label{comparison}
\begin{tabular}{p{0.20\textwidth}|p{0.35\textwidth}|p{0.35\textwidth}}
	\hline
  Characteristic & GrowCut & Modified Fuzzy GrowCut\\ 
	\hline
	Selection of Seeds & Selection of seeds of object class and background class. & Selection of seed only of object class.\\
	\hline
  Initialization & All the seeds have strength value equal to 1. & Only the cell corresponding to the center of mass of points has strength  value equal to 1.\\
	\hline
  Segmentation & Based on knowledge of seeds localization provided by the user. & Based on knowledge of seeds localization and in the Gaussian function that separates the region of foreground and background region. \\
	\hline
  Fault Tolerance to seeds localization & Low & High\\   
	\hline
\end{tabular}
\end{table}

\textbf{The main difference between the classical GrowCut and our proposal is the tolerance to wrong seed positioning. The introduction of Gaussian fuzzy membership functions whose parameters are determined by statistics of the selected points positions makes our proposal more fault-tolerant and, consequently, less dependent on users accurate positioning of seeds. The initialization based on the center of mass also contributes to the method be more flexible in relation to a wrong seed positioning. Furthermore, our proposal reduces of the effort to select seeds, once our approach requires users to choose just the points internal to the object of interest, because the background region is determined by the complement of the Gaussian fuzzy membership function responsible to regulate the strength and the label of each cell at the updating process.}

\begin{figure}[!htb]
\centering
\includegraphics[width=\textwidth]{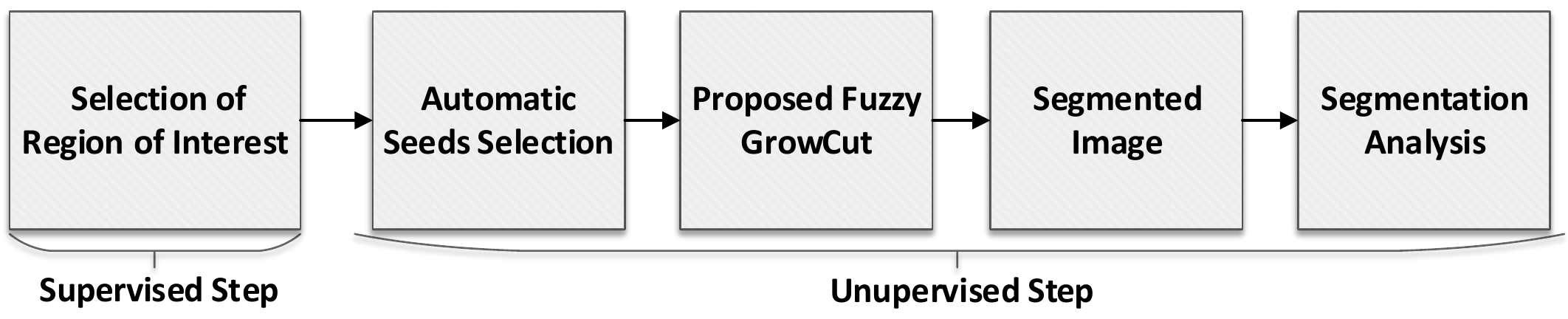}
\caption{Flowchart of the proposed semi-supervised methodology, using Fuzzy GrowCut.}
\label{fig:flowchart}
\end{figure}

The flowchart of Figure \ref{fig:flowchart} illustrates the methodology we are proposing. Firstly, a specialist selects the region of interest, generating a sub-image as input for the automated process. Afterwards, the Differential Evolution optimization algorithm automatically selects points internal to the probable mammary lesion. These seed points are located by an optimization process guided by a fitness function to maximize the distance between points and the brightness of the pixels positions. Subsequently, our modified GrowCut algorithm performs the segmentation, generating a labeled image. Hence, once given a region of interest, segmentation is performed  automatically. Figure \ref{fig:process} illustrates the steps of the proposed method for some images of the Mini-MIAS database.

The selection of seeds consists of identifying initial pixels located in regions of tumor and non-tumor. In many seed-based techniques, seeds are selected manually by a specialist, such as Random Walks, described by \cite{grady2006random}, and Graph Cut, used by \cite{vicente2008graph}. An important characteristic of the proposed algorithm is that it is not necessary to select non-tumor seeds, because the proposed algorithm can adjust its Gaussian Fuzzy frontier based only on the seeds of the tumor region. 

In this work, we used the Differential Evolution optimization algorithm \cite{Das2010} to find automatically seeds in the region of the object of interest, i.e. the suspicious lesion. The solution is represented as a set of seeds, which explore the coordinates of image to find the best representations of initial seeds. The number of seeds per solution is an algorithm parameter which is configured in the adaptive step of the proposed method. Knowing that mass regions frequently have higher intensity pixel values, the problem of finding an adequate set of seeds was converted into an optimization problem, in which we are interested in finding a set of points with maximum distance between each one, and maximum sum of associated gray levels, in order to obtain a non-collapsing set of points located inside the brighter areas. For this purpose, we used a multiobjective fitness function, which evaluates both the distance between seed points and the levels of intensity of the pixels related to them.

\subsection{Experimental Environment} 

The proposed work used 40 images from the Mini-MIAS database, developed by \cite{suckling1994mammographic}. Mini-MIAS images are 200$\mu m$ pixel edge with resolution of 1024 $\times$ 1024 pixels. The ground truth image, which is the expected segmented image, was obtained by a pre-built supervised auxiliary software, based on an adaptive threshold, which uses the indications provided from the database. Among the selected images there are mammograms with different shapes of tumor: circumscribed, spiculated and indistinct.

	The approach used in this work is based on a system which will receive as input the regions of interest previously selected by a specialist. Once the region of interested in obtained, it is submitted as input to the segmentation algorithms. Most of techniques which found a region of interest do not provide analysis about the quality of segmentation in terms of shape of the tumor. Once the contour quality of segmentation is important to define the type of tumor, is important to have better algorithms to enhance the final contour of the segmentation. Furthermore, several recent works are also based on regions of interest provided by a specialist, as used by \cite{esener2015new} and \cite{sharma2013roi} \cite{sharma2015computer}. Some databases, as the IRMA, described by \cite{oliveira2008toward}, is composed only by ROIs, called patches, which include ROIs of mini-MIAS.

\subsection{Metrics} \label{sec:metrics}

T\textbf{he analysis of results performed is based on 10 metrics of analysis: Area, Perimeter, Form Factor, Solidity, Feret's Distance, Dice Similarity Coefficient, Sensitivity, Specificity, Balanced Accuracy and Slope Pattern. We used metrics related to the contour and form because we wanted to evaluate the quality of contour and form of segmented region of the proposed techniques, once size and shape of masses determine the classification in benignant or malignant. }Each one of the metrics is described as following.

The Area corresponds to the number of pixels in the region occupied by the object of interest, i.e. the suspicious lesion. It is expected that segmented images and ground truth images have approximately the same numerical areas.

The Perimeter is the distance, in pixels, around the boundary of the region. Although regions with different shapes can have the same perimeter, when combined with other metrics, it is important to evaluate the similarity of regions.

The Form Factor furnishes a comparison between the area of a polygon with the square of its perimeter. In a circle, the Form Factor is 1, whilst we have $\frac{\pi}{4}$ for a square. This metric is defined by the expression of Equation \ref{eq:ff}:
\begin{equation}
\mathrm{FormFactor} = \frac{4 \pi \mathrm{Area}}{\mathrm{Perimeter}^2}.
\label{eq:ff}
\end{equation}

Solidity is a measure of the intersection between the proper area and the convex area, given by the expression of Equation \ref{eq:solidity}:
\begin{equation}
\mathrm{Solidity} = \frac{\mathrm{Area}}{\mathrm{ConvexArea}}.
\label{eq:solidity}
\end{equation}

The solidity value of a convex polygon, with no holes, is equal to 1.0, representing a solid object. When solidity is lower than 1.0, the object is considered irregular or contains a hole. Hence, solidity helps to indicate the shape of the object.

Feret's diameter is the longest distance between any two points along the selection boundary, also known as maximum caliper. In this work, the angle of Feret's diameter is calculated for 0 and 90 degrees, which represents the maximum distance in coordinates $x$ and $y$, named as Feret X (FX) and Feret Y (FY). It is important to note that the metrics related to shape should be analyzed together, not individually.

{\bfseries 

The Dice Similarity Coefficient (DSC), first proposed by \cite{dice1945measures}, is a spatial overlap index and a reproducibility validation and has been used as validation of a image segmentation quality \citep{zou2004statistical}. The value of a DSC ranges from 0, indicating no spatial overlap between two sets of binary segmentation results, to 1, indicating complete overlap. The DSC measures the spatial overlap between two segmentations, A and B target regions, and is defined as  by Equation \ref{eq:dsc}:

\begin{equation}
\mathrm{DSC} =\frac{2\cdot \left | \mathrm{Seg} \cap \mathrm{GT} \right |}{\left | \mathrm{Seg} \right | + \left | \mathrm{GT} \right |},
\label{eq:dsc}
\end{equation}

\noindent where $\mathrm{Seg}$ represents the segmented image and $\mathrm{GT}$ the ground-truth.

Sensitivity measures the true positivity recognition rate. In this work, sensitivity is evaluated for the classification of pixels, being described in equation \ref{eq:sensivity}.
\begin{equation}
\mathrm{Sensitivity} = \frac{ \left | F_\mathrm{Seg}\bigcap F_\mathrm{GT} \right |}{ \left | F_\mathrm{Seg}\bigcap F_\mathrm{GT} \right | + \left | B_\mathrm{Seg} \bigcap F_\mathrm{GT} \right |},
\label{eq:sensivity}
\end{equation}

\noindent where $F_\mathrm{Seg}$ and $F_\mathrm{GT}$ are the number of foreground pixels of the segmentation and ground-truth, respectively, whilst $B_\mathrm{Seg}$ and $B_\mathrm{GT}$ are the number of background pixels of segmentation and ground-truth, in this order.

Specificity measures the true negative recognition rate. It is described in equation \ref{eq:specifity}.
\begin{equation}
\mathrm{Specificity} = \frac{\left | B_\mathrm{Seg}\bigcap B_\mathrm{GT} \right |}{\left | B_\mathrm{Seg}\bigcap B_\mathrm{GT} \right | + \left | F_\mathrm{Seg} \bigcap B_\mathrm{GT} \right |}.
\label{eq:specifity}
\end{equation}

Having a high sensitivity and low specificity does not necessarily mean good segmentation. Therefore, the Balanced Accuracy (BAC) is defined as the average of Sensitivity and Specificity, describing both the negative and positive recognition rates, as described in equation \ref{eq:balancedAc}.
\begin{equation}
\mathrm{BAC} = \frac{\mathrm{Sensitivity} + \mathrm{Specificity}}{2}.
\label{eq:balancedAc}
\end{equation}

The Slope Spectrum Pattern (SSP), proposed by \cite{toudjeu2008global}, illustrates the shape of the tumor through a signature of an image. If two images have similar Spectrum Pattern, they are considered similar. The Slope Pattern
Spectra algorithm extracts increasing slope segments as pattern spectra. A slope segment is defined as the variation in terms of intensity value when moving from one pixel location to another. This variation can also be referred to as a discrete derivative function. In the proposed work, the SSP of segmented images are compared with the SSP of the ground truth images to evaluate the similarity of shape between them.

%

}

\subsection{Comparison with segmentation techniques}

For the complete evaluation of GrowCut, efficient segmentation techniques in the area were selected and experiments are shown in section \ref{sec:results}. For the comparison we selected the following techniques: Active Contours, Region Growing, Random Walks and Graph Cut, which are representative techniques in the area of segmentation.

Active Contours, proposed by \cite{chan2001active}, is a technique extensively used to identify contours of images. It attempts to minimize the energy associated with the current contour as a sum of internal and external energy. It needs an initial user interaction to select a region which contains the object of interest and the algorithm continues iterating until it identifies a final contour. Some research has been done applying Active Contours to mass segmentation, with some adaptations, as performed by \cite{tunali2013mass} and \cite{rahmati2009maximum}. \textbf{In this work we used Active Contours as proposed by \cite{chan2001active} and implemented by \cite{code_activecontours}.}

Region Growing is a classical image processing technique, which has been applied to segment images in several areas, including mammography, as proposed by \cite{chakraborty2012detection}. Region Growing is a simple segmentation technique, in which the user initially selects seed points and the algorithm adds neighbor pixels that are similar to the seeds. When the region stops growing the algorithm converges and a final segmentation is obtained. \textbf{For this algorithm we used a self-made code.}

Random Walks, as described by \cite{grady2006random}, is a graph-based method, in which each pixel is represented as a node and it is connected to neighbor pixels by edges. The weighted graph starts with a definition of background and object seed points, defined by the user. This technique has also been used by \cite{zhengrandom} to segment masses in mammograms, with successful results. \textbf{The implementation of Random Walks used in this work is used by\cite{andrews2010fast} and provided by \cite{code_randomwalks}}.

Finally, Graph Cut is another graph-based method often used for image segmentation. It requires the selection of seed pixels, as containing the object or the background, and the minimum cut of the graph will determine the energy function to be minimized locally or globally. In mass segmentation of mammograms, it was applied by \cite{wu2011interactive}, which obtained good results when combined with the Watershed technique, described by \cite{beucher1993morphological}. \textbf{The implementation of Graph Cut used in this work is the same as proposed by \cite{boykov2004experimental}, whose code was implemented by \cite{Bagon2006} .
}

\section{Results} \label{sec:results}

The results section was divided into four parts: one for canonical GrowCut results, showing the segmentation based on user interaction, one section of Semi-Supervised GrowCut, showing the results of this technique, \textbf{one section showing the results of the Fuzzy GrowCut and the fourth section with comparative results between both versions of GrowCut and related techniques}.

\subsection{Results of Supervised GrowCut}

\begin{figure}[htp]
\centering
 \includegraphics[width=7cm]{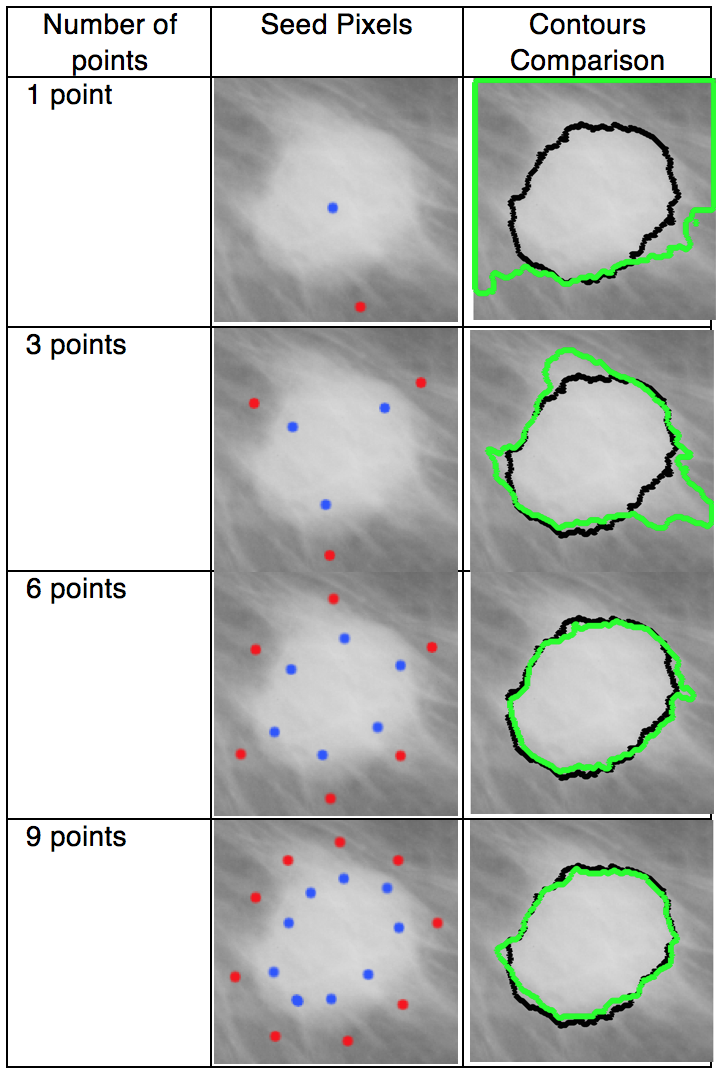}
  \caption{Comparison of the quality of segmentation based on the number of seed pixels, to image mdb028}
  \label{fig:comp_growcut}
\end{figure}

In order to evaluate the performance of the Grow Cut technique, a set of 40 images from the Mini-MIAS database were selected for analysis. The first analysis was performed to evaluate the behavior of the relation between the number of seed pixels in the GrowCut technique and the impact on the quality of segmentation. For this purpose, four cases were observed: selection of 1, 3, 6 and 9 seed points, for each class of segmentation. Figure \ref{fig:comp_growcut} shows the results of this experiment for image mdb028.

The first column of the table shows the number of points for each class: tumor and non-tumor. The second column shows the selected points, where the red points are the points external to the tumor and the blue ones are the points inside. The third column shows a comparison between the contour segmentation obtained through the application of the GrowCut technique, represented by the green line, and the ground truth contour, represented by the black line.

\begin{figure}[htp]
\centering
 \includegraphics[width=7cm]{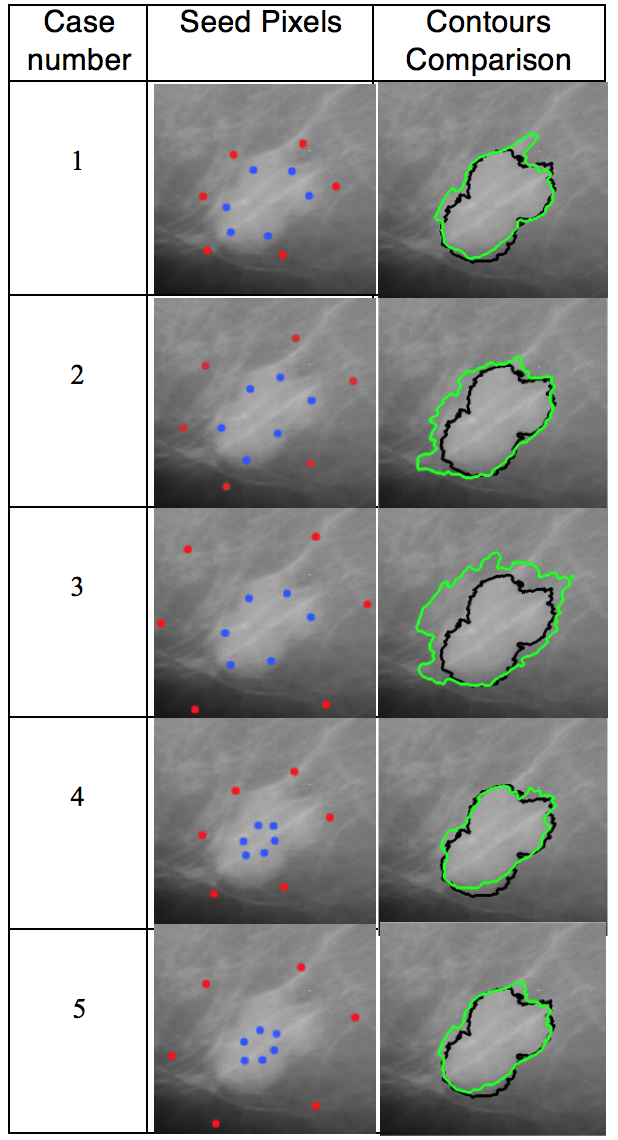}
  \caption{Analysis of distance of seed pixels to the tumor edge, for image mdb005}
  \label{fig:comp_dist}
\end{figure}

The second analysis was intended to observe the impact of the localization of the points on the quality of the segmentation. This analysis is shown in Figure \ref{fig:comp_dist}. The first scenario was done placing the internal and external points close to the edges of the tumor. As expected, it provides a good segmentation with this selection. In case 2 the external points are put not so close to the tumor edges. The result is that the segmentation has its quality decreased, but it is still close to the ground truth contour. If the external points are put further away, as in case 3, the accuracy of the technique is not so good. In cases 4 and 5, the distance of internal points are changed, while maintaining the external points closer and at middle distance, respectively.

\begin{figure}[htp]
\centering
 \includegraphics[width=13cm]{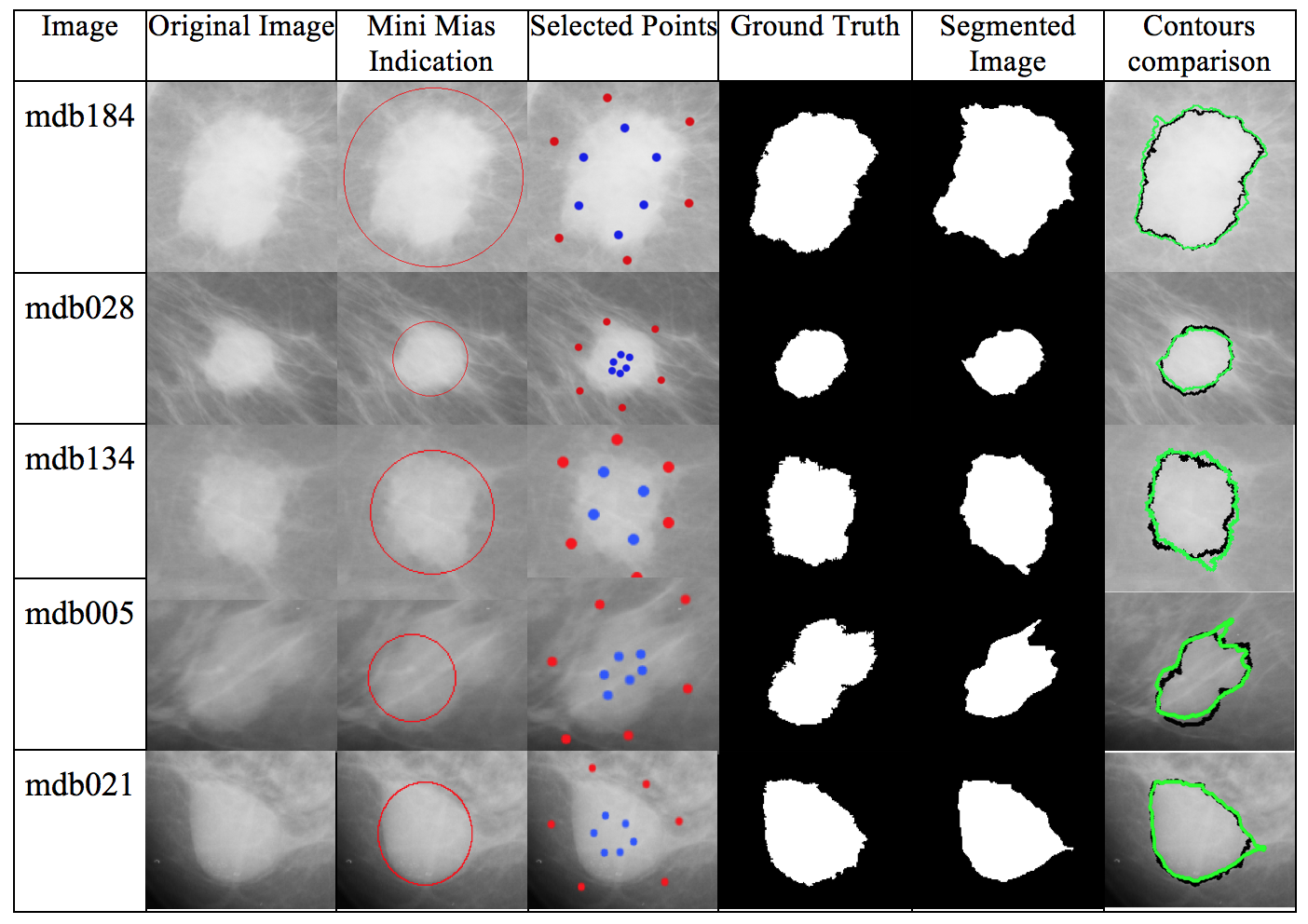}
  \caption{Comparison between GrowCut segmentation and ground truth}
  \label{fig:gc_seg}
\end{figure}

The next analysis was done comparing the GrowCut segmentation to the ground truth. Figure \ref{fig:gc_seg} illustrates 5 cases of well-performed segmentation. \textbf{The first column of the Figure \ref{fig:gc_seg} shows the name of image in the Mini-MIAS. The second column of the figure provides the region of interest of the original image, in the Mini-MIAS database. The third column shows the tumor indication of the database, represented as a red circle, while the fourth column shows the points selected for the application of the GrowCut technique.}
 
 In GrowCut, it is necessary to select points inside and outside of the expected segmentation area. In the third column of Figure \ref{fig:gc_seg}, the outside points are marked in red, while the inside points are marked in blue. On average, it was necessary to select around 6 outside points and 5 inside points, which equal a total of 11 points. However, the number of points can vary depending on the complexity of segmentation. \textbf{Selecting more points makes the segmentation more accurate, but requires more effort selecting them. In the fifth column of Figure \ref{fig:gc_seg} the ground truth of the tumor segmentation is shown. This ground truth was constructed finding the best threshold for each image. Next, the sixth column represents the segmentation result of the GrowCut technique.} The segmentation is done automatically once provided the points of the fourth column. Finally, the last column represents the comparison between the contours of the GrowCut segmentation, in the green line, and the ground truth contour, in the black line. The cases analyzed so far are also evaluated statistically comparing metrics of area, perimeter, form factor, Feret's distance and solidity.

For each metric, an estimated error is calculated, comparing the metric for the segmented image and the ground truth. The estimated error is determined by Equation \ref{eq:ErrorMetric}:

\begin{equation}
\mathrm{Error}_\mathrm{Metric}=\left| 1-\frac{\mathrm{Metric}_\mathrm{seg}}{\mathrm{Metric}_\mathrm{gt}} \right|,
\label{eq:ErrorMetric}
\end{equation}

\noindent where $\mathrm{Error}_\mathrm{Metric}$ represents the error of a determined metric, $\mathrm{Metric}_\mathrm{seg}$ represents the metric applied to the segmented image, and $\mathrm{Metric}_\mathrm{gt}$ is the metric applied to the ground truth image. In the estimated error, the closer the error is to zero the better the result of the segmentation. A low error value represents a high similarity between the segmentation and the ground truth, considering the metric analyzed. Table \ref{tab:metrics_gc} shows the results of estimated errors of the images presented in Figure \ref{fig:gc_seg}.

\begin{table}[htb]
\caption{Estimated errors for the metrics of the form factor, area, perimeter, Feret's distance and solidity}
\centering
\begin{tabular}{l|lllll}
\cline{1-6}
Metric      & mdb184 & mdb028 & mdb134 & mdb005 & mdb021 \\ 
\hline
Form Factor & 0.298  & 0.170  & 0.371  & 0.270  & 0.107  \\
Area        & 0.098  & 0.077  & 0.059  & 0.162  & 0.034  \\
Perimeter   & 0.093  & 0.045  & 0.138  & 0.053  & 0.035  \\
FeretX      & 0.025  & 0.144  & 0.014  & 0.223  & 0.021  \\
FeretY      & 0.082  & 0.019  & 0.067  & 0.107  & 0.000  \\
Solidity    & 0.003 & 0.009 & 0.019  & 0.017  & 0.010  \\ 
\hline
\end{tabular} \label{tab:metrics_gc}
\end{table}

The results of average, maximum, minimum, and standard deviation values observed for the whole set of images analyzed, for the metrics of Form Factor, Area, Perimeter, FeretX FeretY and Solidity are shown in Table \ref{tab:av_metrics}.

\begin{table}[htb]
\caption{Results of average, maximum, minimum, and standard deviation error for the metrics of Form Factor, Area, Perimeter, FereteX, FeretY and Solidity, for all analyzed images}
\centering
\begin{tabular}{l|llll}
\hline
Metric      & Average & Max   & Min   & Std. Dev. \\ 
\hline
Form Factor & 0.272   & 0.968 & 0.001 & 0.215     \\
Area        & 0.139   & 0,713 & 0.014 & 0.145     \\
Perimeter   & 0.152   & 0.486 & 0.017 & 0.112     \\
FeretX      & 0.132   & 0.662 & 0.013 & 0.137     \\
FeretY      & 0.129   & 0.646 & 0.000 & 0.130     \\
Solidity    & 0.071   & 0.261 & 0.001 & 0.067     \\ 
\hline
\end{tabular}
    \label{tab:av_metrics}
\end{table}

Table \ref{tb:gc_avg_metrics2} shows the average values for obtained for the metrics of DSC, Sensitivity, Specificity and Balanced Accuracy.

\begin{table}[htb]
\centering
\caption{Results of average, maximum, minimum, and standard deviation error for the metrics of DSC, Sensitivity, Specificity and BAC, for all analyzed images}
\label{tb:gc_avg_metrics2}
\begin{tabular}{lllll}
\hline
Metric      & Average & Max   & Min   & Std. Dev. \\
\hline
DSC         & 0.861   & 0.969 & 0.616 & 0.084     \\
Sensitivity & 0.901   & 1.000 & 0.731 & 0.084     \\
Specificity & 0.944   & 0.043 & 0.999 & 0.828     \\
BAC         & 0.923   & 0.984 & 0.810 & 0.045    \\
\hline
\end{tabular}
\end{table}

\subsection{Results of Semi-Supervised GrowCut}

The analysis of SSGC was applied to the same set of images used in the GrowCut technique. \textbf{Initially, an empirical experimentation was performed to identify the best multi-layer threshold parameters to the data. The analysis was performed observing the value of metric DSC. The results are shown in Table \ref{tb:par_mlt}}.

\begin{table}[]
\centering
\caption{Analysis of Multi-layer threshold parameters, based on DSC metric.}
\label{tb:par_mlt}
\begin{tabular}{lll}
\hline
Level & Depth & DSC       \\
\hline
\textbf{10}    & \textbf{2}     & \textbf{0.64$\pm$0.21} \\
10    & 3     & 0.64$\pm$0.22 \\
10    & 4     & 0.60$\pm$0.23 \\
10    & 5     & 0.00$\pm$0.00 \\
10    & 6     & 0.00$\pm$0.00 \\
\hline
20    & 2     & 0.53$\pm$0.26 \\
20    & 3     & 0.61$\pm$0.24 \\
20    & 4     & 0.64$\pm$0.22 \\
20    & 5     & 0.64$\pm$0.22 \\
20    & 6     & 0.63$\pm$0.21 \\
\hline
30    & 2     & 0.44$\pm$0.24 \\
30    & 3     & 0.52$\pm$0.26 \\
30    & 4     & 0.56$\pm$0.26 \\
30    & 5     & 0.62$\pm$0.23 \\
30    & 6     & 0.63$\pm$0.22 \\
\hline
40    & 2     & 0.37$\pm$0.23 \\
40    & 3     & 0.46$\pm$0.24 \\
40    & 4     & 0.54$\pm$0.24 \\
40    & 5     & 0.56$\pm$0.26 \\
40    & 6     & 0.59$\pm$0.26 \\
\hline
50    & 2     & 0.33$\pm$0.22 \\
50    & 3     & 0.41$\pm$0.23 \\
50    & 4     & 0.48$\pm$0.24 \\
50    & 5     & 0.52$\pm$0.26 \\
50    & 6     & 0.54$\pm$0.26\\
\hline
\end{tabular}
\end{table}

\textbf{Column 1 and 2 of Table \ref{tb:par_mlt} represents the values of parameters of multi-layer threshold approach, and the third column is the value of the metric DSC. Some pair of configurations obtained close values, but we chose the parameters of the first line of Table \ref{tb:par_mlt}, because it had the higher value of DSC with the lower standard deviation.}

\begin{figure}[htb]
\centering
 \includegraphics[width=15.5cm]{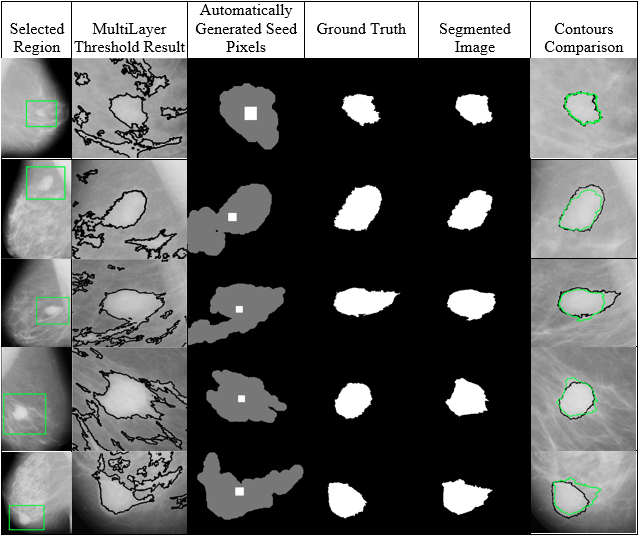}
  \caption{Comparison between Semi-Supervised GrowCut segmentation and ground truth}
  \label{fig:ssgc_seg}
\end{figure}

A qualitative and a quantitative analysis was done to observe the performance of segmentation using SSGC. The qualitative analysis is shown in Figure \ref{fig:ssgc_seg}, where it is divided into \textbf{6 columns}, each one showing one step of the process, where each row corresponds to a different image. The first column shows the region that the user selected from the original image, from the Mini-MIAS database. The SSGC initially requires the manual selection of the ROI to perform well, but it can be extended to be fully unsupervised. Columns 2 to 6 show only the region of interest selected by the user, not the whole image. The images from column 2 represent the segmentation results from multilevel threshold. As can be seen, it does not provide a perfect segmentation, but it is useful to select the seeds automatically.

From the results showing the multilayer threshold, only the region with the bigger area is maintained to remove some undesirable extensions and islands. After that, a dilation is performed to identify the seed pixels external to the tumor and a centroid calculus to identify the internal seed pixels. In the third column, the area in black represent the seeds external to the tumor, which are the pixels classified with the label non-tumor. The area in white represents the pixels that are identified as internal to the tumor, which are the pixels with the label tumor. The area in gray shows the pixels which are not yet labeled and that will be labeled after the application of GrowCut. Those labeled pixels serve as entry to the GrowCut mechanism. Columns 4 and 5 show the ground truth and SSGC segmentation, respectively. The last column shows a comparison between the contours of the ground truth, in the black line, and SSGC segmentation, in the green line. 

\begin{figure}[htb]
\centering
 \includegraphics[width=7.5cm]{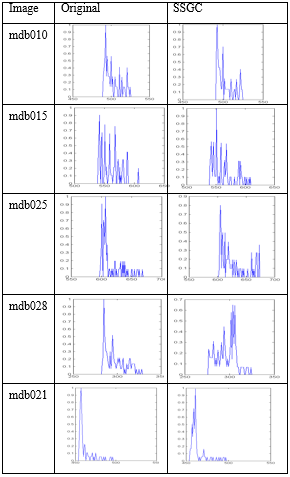}
  \caption{Comparison of the Slope Spectrum Pattern of ground truth and Semi-Supervised GrowCut.}
  \label{fig:spectro}
\end{figure}

Other analysis was performed applying the Slope Spectrum Pattern. This analysis was performed on all images and in most cases segmented images presented a similar Spectrum Pattern to the ground truth segmentation. The Slope Patterns of the segmentation results in Figure \ref{fig:ssgc_seg} are presented in Figure \ref{fig:spectro}. 

Finally, the Wilcoxon Signed Ranked Test was applied to each pair of segmented and ground truth images, comparing the Slope Spectrum Pattern distribution. The test was applied to the whole set, using a confidence value of 0.05. The null hypothesis was that each pair of samples was equal. A total of 77.5\% of images passed on the test and did not reject the null hypothesis that segmented image and ground truth were equal. The results, including p-value, min, max and standard deviation are shown in Table \ref{tab:wilco}.

\begin{table}[htb]
\caption{Wilcoxon Signed Rank Test applied to segmented and ground truth images, comparing the Slope Spectrum Pattern distribution, for the SSGC algorithm.}
\centering
\begin{tabular}{cccccc}
\hline
\begin{tabular}[c]{@{}c@{}}Average \\ p-value\end{tabular} & \begin{tabular}[c]{@{}c@{}}\#Reject Null \\ Hypotheses\end{tabular} & \begin{tabular}[c]{@{}c@{}}\#Not Reject \\ Null Hypotheses\end{tabular} & Min p-value  & \begin{tabular}[c]{@{}c@{}}Max p \\ Value\end{tabular} & \begin{tabular}[c]{@{}c@{}}Standard \\ Deviation\end{tabular} \\ \hline
0.3632                                                     & 9                                                                   & 31                                                                       & 8.67 e-04 & 0.9875                                                 & 0.3047                                                        \\ \hline
\end{tabular}
    \label{tab:wilco}
\end{table}

\subsection{Results of Fuzzy GrowCut}

\begin{figure}[!htb]
\centering
\includegraphics[width=9cm]{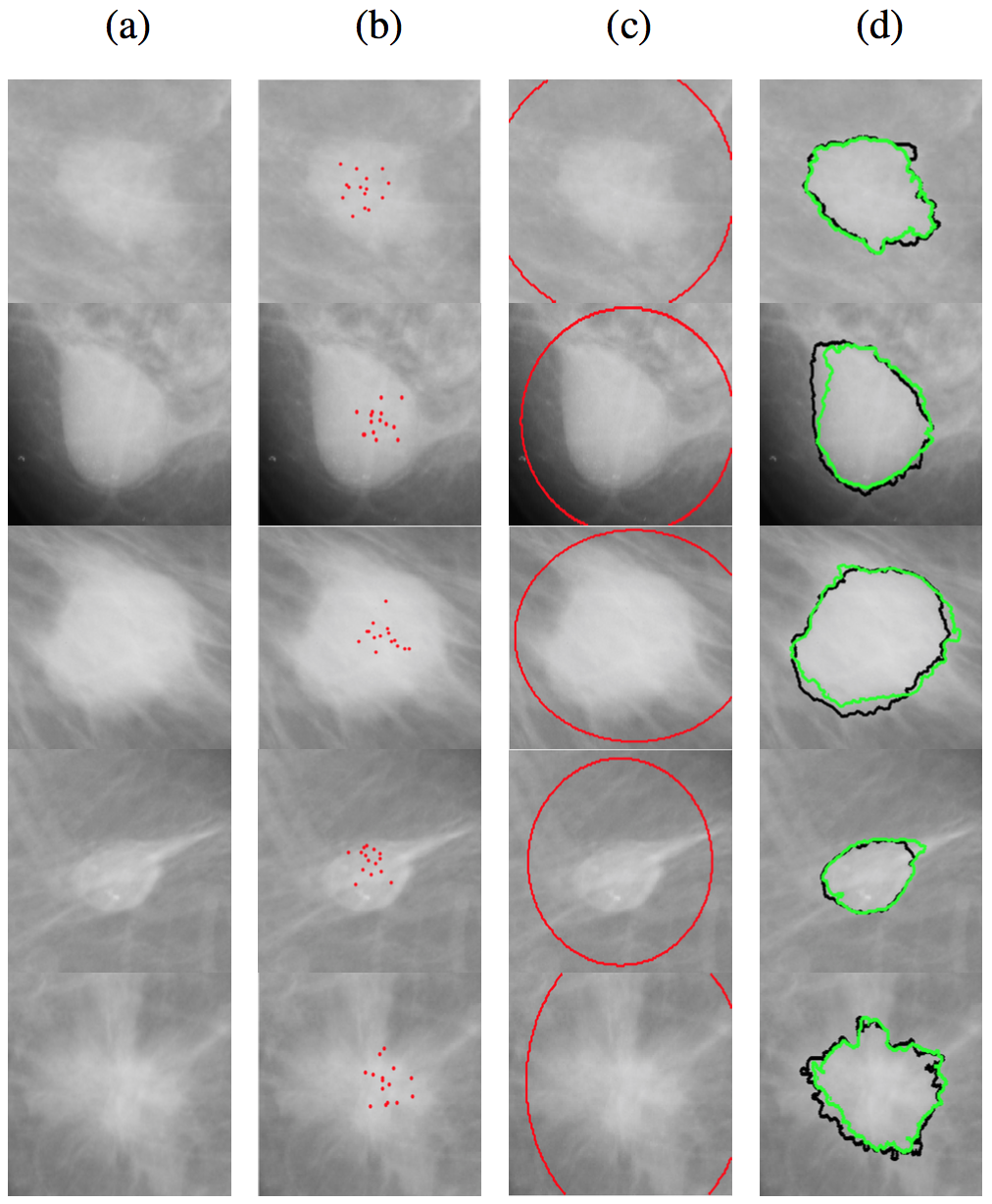}
\caption{Images mdb010, mdb021, mdb028, mdb132, mdb175, and mdb181, obtained from Mini-MIAS database, illustrating the proposed semi-supervised segmentation method. (a) Original Image; (b) Automatic generated seeds; (c) Gaussian fuzzy region; (d) final segmentation (green) compared to ground truth image (black).} \label{fig:process}
\end{figure}

The proposed Fuzzy GrowGut (FGC) was evaluated using the 40 images of mini-MIAS database, using the metrics described in section \ref{sec:metrics}. The semi-supervised Fuzzy GrowCut approach uses two parameters: number of points of a solution in Differential Evolution Algorithm, used to automatically generate the seeds, and the $\alpha$ value defined in Equation \ref{eq:probobj}. The parameters were empirically tuned 30 and 2, for the number of points and $\alpha$ value, respectively. Figure \ref{fig:process} shows the results segmentation to the images mdb010, mdb021, mdb028, mdb132, mdb175, and mdb181.

Column \emph{a} of Figure \ref{fig:process}	 represents the initial region of interest, manually selected from the Mini-MIAS database. Column \emph{b} shows the seed points, in red, obtained from automatic seeds selection. As cited in  section \ref{sec:fgc}, the proposed technique requires only seeds of the tumor region, which is an advantage compared to other techniques. Column \emph{c} illustrates the boundaries of the Gaussian fuzzy region, where the inside region has higher membership degrees of pixels internal to the lesion mass. The size of the Gaussian region is based on the location of seed points. Finally, column \emph{d} shows the final segmentation of the proposed approach, in green, compared with the ground-truth contour, in black.

\subsection{Comparison with related techniques}

A comparative study was performed between supervised GrowCut, Semi-Supervised GrowCut (SSGC), Fuzzy GrowCut (FGC), Region Growing, Random Walks, Active Contours, and Graph Cut. This analysis was performed on the same dataset as the previous experiments, observing the quality of final segmentation of each technique and the provided ground truth.

\begin{figure}[htb]
\centering
 \includegraphics[width=15.0cm]{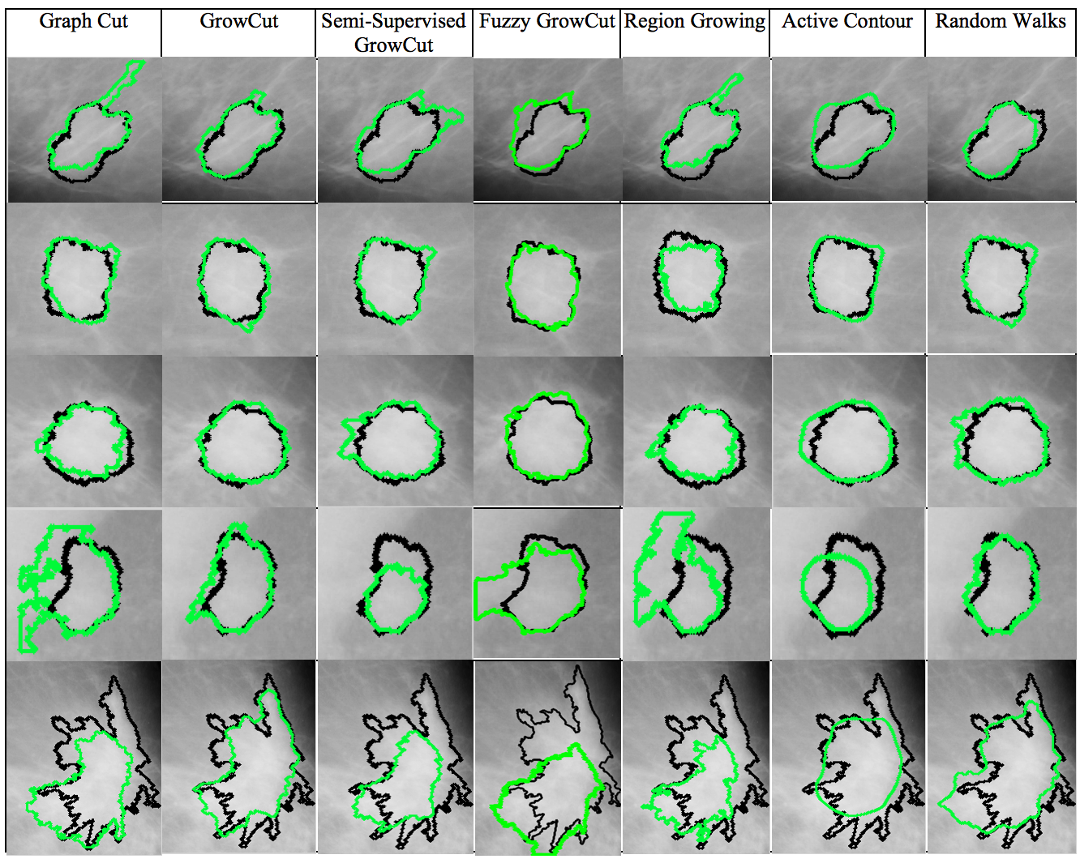}
  \caption{Comparison of final segmentation between Graph Cut, GrowCut, Semi-Supervised GrowCut, Region Growing, Active Contour, and Random Walks}
  \label{fig:todas}
\end{figure}

Figure \ref{fig:todas} illustrates some cases of final segmentation of each technique. Each column of \textbf{Figure \ref{fig:todas}} shows the result of segmentation of each technique for a selected region of interest, and each row corresponds to a different image from Mini-MIAS. In each cell, the black contour of the image represents the ground truth contour, while the green one corresponds to the final segmentation of the technique.
 
\begin{figure}[htb]
\centering
 \includegraphics[width=16.0cm]{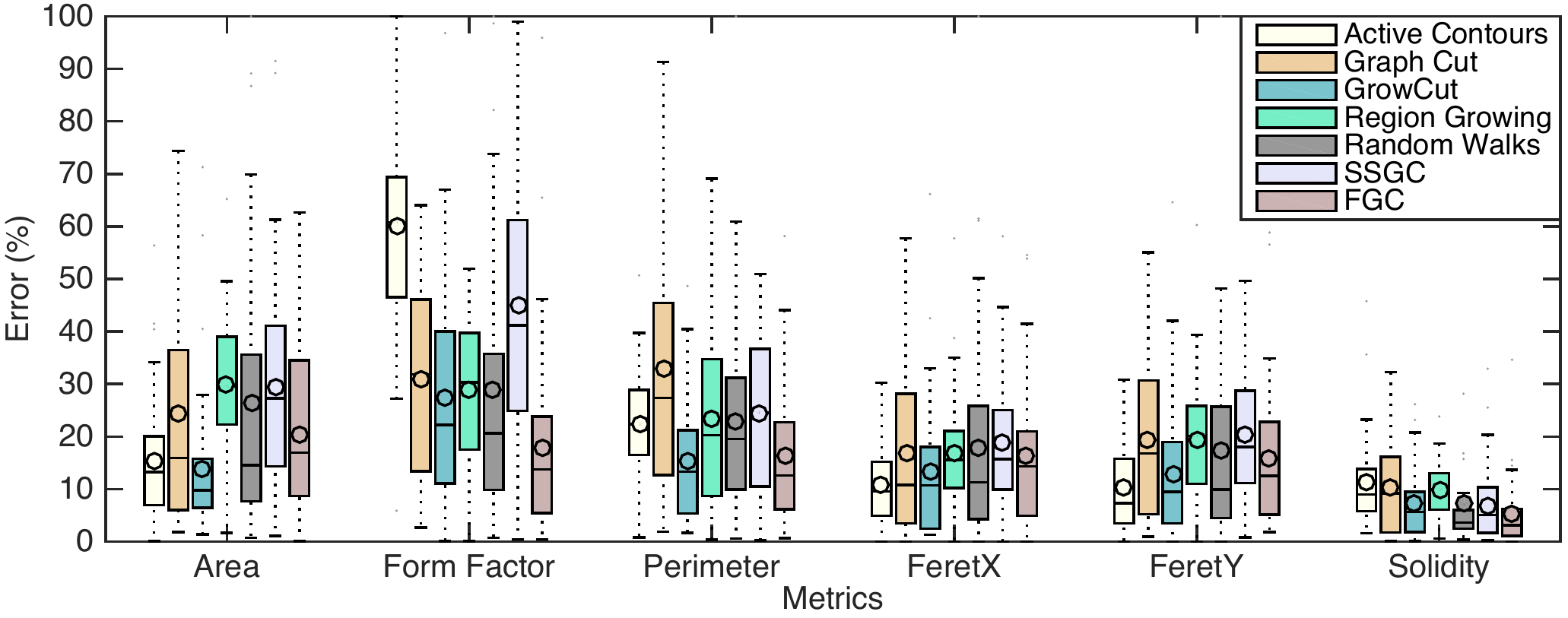}
  \caption{Average error of Active Contours, Graph Cut, GrowCut, Region Growing, Random Walks, SSGC and Fuzzy GrowCut to the metrics of Area, Form Factor, Perimeter, FeretX, FeretY and Solidity}
  \label{fig:boxplot}
\end{figure}

The last study was performed observing the metrics described previously. The metrics were applied for each technique and obtained the average relative error of the segmentation, as described in Equation \ref{eq:ErrorMetric}. The relative error is estimated to the metrics of area, perimeter, Feret's distance and solidity, for all techniques and images analyzed. The lower the relative error, the better is the quality of segmentation. Figure \ref{fig:boxplot} shows relative error, in percentage, with the boxplot analysis for these metrics. The lower and smaller the boxplot, the better is the result.

\begin{figure}[htb]
\centering
 \includegraphics[width=15.0cm]{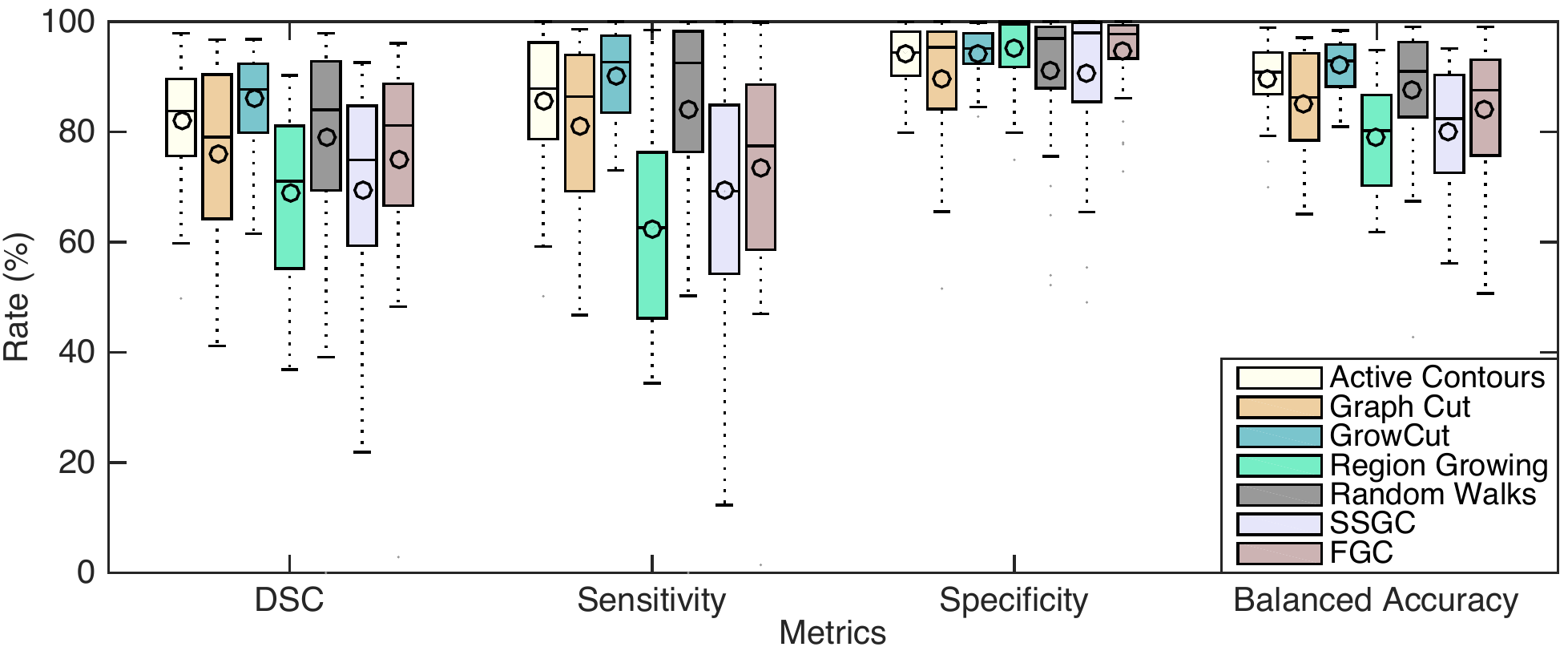}
  \caption{Average results of Active Contours, Graph Cut, GrowCut, Region Growing, Random Walks, SSGC and Fuzzy GrowCut to the metrics of DSC, Sensitivity, Specificity and Balanced Accuracy}
  \label{fig:boxplot_p2}
\end{figure}

\textbf{Figure \ref{fig:boxplot_p2} shows the results obtained for the metrics DSC, Sensitivity, Specificity and Balanced Accuracy. For these metrics, the higher the  value the better is the result.}

\textbf{Finally, we performed the Wilcoxon Signed Ranked Test applied to each pair of segmented and ground truth images, comparing the Slope Spectrum Pattern distribution. The test was applied to the whole set, using a confidence value of 0.05. The null hypothesis was that each pair of samples was equal. The results, including p-value, min, max and standard deviation are shown in Table \ref{tab:slope_all}. Having a rejected null hypothesis means that the Slope Spectrum Pattern distribution of segmented image and ground truth are different. That means that the shape of segmented and ground truth images have significative difference. From \ref{tab:slope_all}, having a higher average p-vale and a lower number of rejected null hypothesis means a better quality of segmentation. 
}

\begin{table}[htb]
\centering
\caption{Wilcoxon Signed Rank Test applied to segmented and ground truth images, comparing the Slope Spectrum Pattern distribution, for the all the techniques analysed.}
\label{tab:slope_all}
\begin{tabular}{ccccccc}
\hline
\begin{tabular}[c]{@{}c@{}}   \\ Techniques \end{tabular} & \begin{tabular}[c]{@{}c@{}}Average \\ p-value\end{tabular} & \begin{tabular}[c]{@{}c@{}}\#Reject Null \\ Hypotheses\end{tabular} & \begin{tabular}[c]{@{}c@{}}\#Not Reject \\ Null Hypotheses\end{tabular} & Min p-value  & \begin{tabular}[c]{@{}c@{}}Max p \\ Value\end{tabular} & \begin{tabular}[c]{@{}c@{}}Standard \\ Deviation\end{tabular} \\ \hline

Active Contours & 0.0491           & 34                         & 6                              & 8.71E-01    & 0.7773      & 0.1447             \\
GraphCut        & 0.4573           & 3                          & 37                             & 7.46E-04    & 0.9877      & 0.3386             \\
GrowCut         & 0.4768           & 4                          & 36                             & 1.10E-03    & 0.9612      & 0.2873             \\
Region Growing  & 0.1443           & 29                         & 11                             & 2.61E-07    & 0.7833      & 0.2476             \\
RW              & 0.4427           & 7                          & 33                             & 1.96E-04    & 0.9841      & 0.3298             \\
SSGC            & 0.3632           & 9                          & 31                             & 8.67E-04    & 0.9875      & 0.3047             \\
FGC             & 0.4931           & 4                          & 36                             & 8.61E-04    & 0.9795      & 0.3479       \\
\hline     
\end{tabular}
\end{table}

\section{Discussion} \label{sec:discussion}

In the GrowCut analysis, for the results presented in Figure \ref{fig:comp_growcut}, for the first case, it could be observed that selecting only one point internal and external to the tumor is not enough to provide a good segmentation. When the number of points is increased to 3, as in the second case of the same table, the segmentation gets better, but there are still some contours not well defined. However, with 6 selected points, as in the third case, the segmentation approaches that of the ground truth image. By increasing the number of points to 9, the segmentation improves, but the effort of selecting more points may not be worthy in some cases. In most of the images tested, selecting 6 internal and external points achieved a better tradeoff between good segmentation efforts and selecting points. Based on this study, we used 6 points of selection in the following analysis. However, the number of points can be adjusted depending on the problem.

Based on the results of Figure \ref{fig:comp_dist}, where the impact of the distance of selected points to the edges of the tumor was analyzed, we conclude that the closer the external points are to the edges of the tumor, the better is the GrowCut segmentation. On the other hand, we can conclude that the distance of internal points to the edge does not have so much impact as the distance of external points. In most cases analyzed, where the tumors were small, the internal location of points does not have much influence. However in some cases, such as bigger tumors or in ill-defined border cases, the internal points can have a significant impact. But in most of the cases it is easier to select central internal points of the tumor and external points close to the edge. This selection is enough to achieve a good quality of segmentation, as verified in the analyzed cases.

From Figure \ref{fig:gc_seg}, it could be observed that the contours of the GrowCut's segmentation are quite similar to the ground truth's contours. The same quality of results found in this analysis can be verified in most of the selected images of the database. Also, it could be observed that all the segmented images are inside the region indications of tumor provided by the database. Even on images that are harder to segment, due to low density of the tumor or because of the poor image quality, the technique is able to provide good segmentation. This fact occurs due to the user providing some information of the regions of tumor and non-tumor, through the seed pixels, is enough to the technique segments the region by itself. Although the selection of points for GrowCut's segmentation requires low effort, the selected points need to be close to the expected segmentation area, so the technique can provide a better segmentation.

As could be observed in Table \ref{tab:metrics_gc}, in all cases the errors were close to zero, for the six metrics analyzed. The form factor represents a similarity of the shape of the segmentation, which occurred in all cases, as could be seen in Figure  \ref{fig:gc_seg}. The area, calculated in pixels, shows correspondence between both segmentation and ground truth. The same occurs with the perimeter, Feret's distance and solidity metrics. 

In Table \ref{tab:av_metrics}, it is shown that the average error of the segmentation is low and with a low standard deviation. For all images tested in the database, the estimated error was close to zero, for the six metrics analyzed. This means that the technique is able to achieve a representative similarity in terms of area, perimeter and shape, which indicates that it is a good solution for semi-automatic segmentation. It is important to notice that average error can be even more reduced if the number of selected points of the GrowCut is increased, improving the quality of segmentation. \textbf{Table \ref{tb:gc_avg_metrics2} shows that the GrowCut also had good average values for the metrics of DSC, Sensitivity, Specificity and Balanced Accuracy, which shows that the overlap are between segmentation and ground truth are close. 
}
When analyzing the Semi-Supervised GrowCut, Figure \ref{fig:ssgc_seg} showed that the final segmentation gets close to the ground truth expected contour. For a semi-supervised technique it is quite an interesting result, once it eliminates the user interaction effort necessary in the original GrowCut and most other techniques. However, the quality of segmentation is not as accurate as the supervised GrowCut, but it is expected because SSGC is a semi-supervised technique. This happens because the selection of seeds is done automatically, and may not be close enough to the edges of the tumor.

From Figure \ref{fig:spectro} it could be observed that the Slope Spectrum Patterns between ground truth and segmented image are similar. This result means that the contours of segmented images and ground truth images are similar, which happens in most of the cases. From the results of the Wilcoxon test, in Table \ref{tab:wilco}, it could be statistically proven that the images from ground truth and segmentation of SSGC are similar. However, some cases did not achieve a good segmentation, as could be seen since 7 cases were rejected. 

	In Figure \ref{fig:process} can be observed that the proposed Fuzzy GrowCut can achieve a good segmentation quality, close to the ground truth. The main advantage of the Fuzzy GrowCut is the reduction of the need of selection of non-tumor seeds, which reduces the specialist knowledge in the segmentation process. Furthermore, the internal seeds were generated automatically using the Differential Evolution algorithm 


In Figure \ref{fig:todas}, where some cases of segmentation results are shown for all techniques analyzed, it could be observed that, for most cases, all the techniques achieved good segmentation accuracy. However, in the mammograms where the masses are spiculated or present low density, which represents more difficult cases, some techniques did not achieve a good segmentation.

In general, GrowCut and Random Walks have a better result, mainly in the most difficult images. GrowCut still requires fewer seed points than Random Walks, which reduces the user interaction effort maintaining the quality of segmentation. In terms of performance, Random Walks is a little faster than GrowCut, but GrowCut is still quite fast for segmenting the ROIs tested. The exact average time for each algorithm is out of the scope of this work. The Semi-Supervised GrowCut, despite most images it segmented obtained a good segmentation, in some images it may be more difficult to obtain a close final contour. However, the SSGC, as a semi-supervised technique, obtained good segmentation results besides having the benefit of not requiring any user interaction effort. The Fuzzy GrowCut algorithm, which also uses a semi-supervised approach, obtained a quality of segmentation close to the supervised methods, requiring no user interaction.

 \textbf{Figure \ref{fig:boxplot} showed the average and dispersion for Area, Form Factor, Perimeter, FeretX, FeretY and Solidity metrics through boxplot representation}. According to it, in general, GrowCut had the lowest average segmentation errors and range between maximum and minimal values when compared to the other techniques. \textbf{The Active Contours, despite having a low error of Area, FeretX and FeretY, did not achieve good results for the Form Factor metric. This happened because it was not able to segment spiculated edges. Therefore, although having the size of area close to the ground truth, the shape of segmentation obtained was considered different because of the missing spiculated edges, for most of cases. All techniques had a minimal error close to zero, which corresponded to the images where the tumors were relatively easy to segment. However, the maximum errors indicate that some techniques, such as SSGC and Graph Cut, did not achieve good results for the metrics of area and perimeter for some images with complex edges. To those metrics, GrowCut and the proposed Fuzzy GrowCut had lower average error, which represents a better segmentation, in general, when compared to the other techniques.} The range of error also was lower, which indicates a better segmentation in the most difficult cases. The Semi-Supervised GrowCut, despite not having the lowest average errors, still obtained good segmentation results for most images analyzed. It could be observed that SSGC can be well applied when the tumor is well defined and not very spiculated. However, in most difficult cases, SSGC segmentation obtained high errors. Furthermore, it still had good results on average for a semi-supervised  technique and can be improved in future works with a more accurate selection of seed points. \textbf{The Fuzzy GrowCut had better results when compared with the SSGC, for all metrics analyzed. When compared to the supervised techniques, the proposed Fuzzy GrowCut presented lower average error value, and even better, when compared to Graph Cut and Region Growing, for most of metrics. Although the comparison of a semi-supervised method with supervised ones seems unfair, these results shows that the Fuzzy GrowCut can obtain a segmentation quality close the techniques of the state-of-the-art, with no need of seeds selection. Furthermore, a supervised Fuzzy GrowCut version of the proposed method also can be used, which has similiar performance to GrowCut, but with tolerance to wrong seed positioning.
}

	\textbf{Figure \ref{fig:boxplot_p2} showed the results of the metrics DSC, Sensitivity, Specificity and Balanced Accuracy. For these metrics, the GrowCut obtained higher values, with a lower range, which indicates it had better segmentation even for images with irregular edges. Active Contours also obtained good results for these metrics. However, as described previously, the metrics should be analyzed together, and although the overlapped area is close to the the ground truth, the shape of the segmented image can be different, as demonstrated previously by the form factor metric. Random Walks and Fuzzy GrowCut also obtained high rates for these metrics.}
	
	\textbf{Finally, Table \ref{tab:slope_all} showed the results related to the Slope Spectrum Pattern metric. For this metric, GrahCut, GrowCut and FFCG obtained similar results, with lower rejected null hypotheses, which means that few images were considered different from ground truth. The Active Contours had the higher number of rejected hypotheses, due to its difficult to deal with spiculated edges. This metric reflects the results obtained from Factor Form metric, which analyses the shape of the segmentation. As the shape is an important decision criteria to obtain a correct diagnosis, a segmented image must have a shape close to the real shape of the tumor. For this metric, GrowCut and the modified Fuzzy GrowCut showed to be competitive and efficient to mass segmentation.
}

\section{Conclusion} \label{sec:conclusion}


This work presented a detailed study of application of the GrowCut technique to segment tumors in mammogram images. Results showed that GrowCut could obtain segmentation images similar to ground truth, having similar values for the metrics analised, which signifies a considerable similarity in shape and size. An average estimated error was calculated for each metric, achieving an error close to zero. Consequently, GrowCut can provide results with good segmentation quality associated with low effort in selecting initial seed points.

A semi-supervised version of GrowCut was proposed, which demonstrated to be able to obtain good results of segmentation. Despite the fact that Semi-Supervised GrowCut is not as accurate as the original GrowCut, it has the advantage of not requiring a user interaction effort and, therefore, high levels of user's specialist knowledge.

A semi-supervised Fuzzy modification in GrowCut algorithm was also proposed. Differently from the classical GrowCut, in which users have to select points internal and external to the object of interest to be segmented, the introduction of a Gaussian fuzzy membership function allows user to select just internal points. This modification reduces the effort of seeds selection, once only the foreground seeds are necessary to estimate the region of the lesion. Furthermore, it is robust to wrong seed positioning, which is as feature not seen in seed based segmentation algorithms. Results showed that the Fuzzy GrowCut obtained good results of segmentation, requiring no human interaction.

Comparisons with Active Contours, Region Growing, Random Walks, and Graph Cuts showed that GrowCut is quite competitive with these techniques in the field of mammography image segmentation. Results showed that GrowCut reached lower average error for most metrics analyzed. The results also showed that the Fuzzy GrowCut can be competitive with the supervised segmentation techniques, with the advantage of being a semi-supervised method.

In this work, it could be noticed that GrowCut and its proposed semi-supervised variations are feasible and suitable technique that can be applied to breast tumor segmentation, being very efficient and with a low user interaction effort. 


\section{Conflict of interest}

The authors have no conflict of interest to report.

\section{Acknowledgments}

The authors would like to thank the Brazilian research agency FACEPE (ref. PBPG-0527-1.03/10), for partial financial support of this work.

\bibliographystyle{gCMB}
\bibliography{gCMBguide}

\end{document}